\documentclass[acmsmall]{acmart}

\usepackage{multirow}
\usepackage{amsfonts}
\usepackage{color}
\usepackage{subfigure}
\usepackage{adjustbox,lipsum}
\usepackage{array}
\usepackage{multirow}

\usepackage{algorithm}
\usepackage{algorithmic}
\usepackage{rotating}

\newcommand{\etal}{{\em et al.\,}}       
\newcommand{\eg}{{\em e.g.}}           
\newcommand{\ie}{{\em i.e.}}           

\definecolor{orange}{rgb}{1,0.5,0}
\definecolor{grey}{rgb}{0.5,0.5,0.5}
\definecolor{new_blue}{rgb}{0.3,0.5,0.8}

\newcommand{\PreserveBackslash}[1]{\let\temp=\\#1\let\\=\temp}
\newcolumntype{C}[1]{>{\PreserveBackslash\centering}p{#1}}
\newcolumntype{R}[1]{>{\PreserveBackslash\raggedleft}p{#1}}
\newcolumntype{L}[1]{>{\PreserveBackslash\raggedright}p{#1}}

\AtBeginDocument{%
  \providecommand\BibTeX{{%
    \normalfont B\kern-0.5em{\scshape i\kern-0.25em b}\kern-0.8em\TeX}}}

\setcopyright{acmcopyright}
\acmJournal{TOMM}
\acmYear{2024} \acmVolume{1} \acmNumber{1} \acmArticle{1} \acmMonth{1} \acmPrice{15.00}\acmDOI{10.1145/3419439}

\acmConference{~}
\acmBooktitle{~}
\acmPrice{~}
\acmISBN{~}

\begin{document}

\title{MultiMatch: Multi-task Learning for Semi-supervised Domain Generalization}

\author{Lei Qi}
\affiliation{%
  \institution{School of Computer Science and Engineering (Southeast University), and Key Laboratory of New Generation Artificial Intelligence Technology and Its Interdisciplinary Applications (Southeast University), Ministry of Education}
  \city{Nanjing}
  \country{China}}
\email{qilei@seu.edu.cn}

\author{Hongpeng Yang}
\affiliation{%
  \institution{School of Cyber Science and Engineering, Southeast University}
  \city{Nanjing}
  \country{China}
  }\email{hp\_yang@seu.edu.cn}


\author{Yinghuan Shi}
\affiliation{%
  \institution{The State Key Laboratory for Novel Software Technology, Nanjing University}
 \city{Nanjing}
 \country{China}
 }\email{syh@nju.edu.cn}

\author{Xin Geng}
\affiliation{\institution{School of Computer Science and Engineering (Southeast University), and Key Laboratory of New Generation Artificial Intelligence Technology and Its Interdisciplinary Applications (Southeast University), Ministry of Education}
 \city{Nanjing}
 \country{China}
 }\email{xgeng@seu.edu.cn}

\thanks{The work is supported by NSFC Program (Grants No. 62206052, 62125602, 62076063), Jiangsu Natural Science Foundation Project (Grant No. BK20210224), and the Xplorer Prize. Corresponding author: Xin Geng.}
\renewcommand{\shortauthors}{~}

\begin{abstract}
Domain generalization (DG) aims at learning a model on source domains to well generalize on the unseen target domain. Although it has achieved great success, most of the existing methods require the label information for all training samples in source domains, which is time-consuming and expensive in the real-world application. In this paper, we resort to solving the semi-supervised domain generalization (SSDG) task, where there are a few label information in each source domain. To address the task, we first analyze the theory of multi-domain learning, which highlights that 1) mitigating the impact of domain gap and 2) exploiting all samples to train the model can effectively reduce the generalization error in each source domain so as to improve the quality of pseudo-labels. According to the analysis, we propose MultiMatch, \ie, extending FixMatch to the multi-task learning framework, producing the high-quality pseudo-label for SSDG. To be specific, we consider each training domain as a single task (\ie, local task) and combine all training domains together (\ie, global task) to train an extra task for the unseen test domain. In the multi-task framework, we utilize the independent batch normalization and classifier for each task, which can effectively alleviate the interference from different domains during pseudo-labeling. Also, most of the parameters in the framework are shared, which can be trained by all training samples sufficiently. Moreover, to further boost the pseudo-label accuracy and the model's generalization, we fuse the predictions from the global task and local task during training and testing, respectively. A series of experiments validate the effectiveness of the proposed method, and it outperforms the existing semi-supervised methods and the SSDG method on several benchmark DG datasets.
\end{abstract}




\maketitle

\section{Introduction}
Deep learning has achieved a remarkable success in many application tasks~\cite{DBLP:conf/cvpr/HeZRS16,DBLP:journals/pami/RenHG017,devlin2018bert,DBLP:journals/tmm/LuoJGLLLG20}, such as computer vision and natural language processing. However, when there is the domain shift between the training set and the test set, the typical deep model cannot effectively work on the test dataset~\cite{DBLP:conf/ijcai/0001LLOQ21,zhou2021domain}. Hence, it requires training a new model for a new scenario, which is infeasible in real-world applications. Recently, a new task named domain generalization (DG) is proposed~\cite{DBLP:conf/icml/MuandetBS13,liu2022category}, where there are several available source domains during training, and the test set is unknown, thus the training and test samples are various from the data-distribution perspective. The goal of DG is to train a generalizable model for the unseen target domain using several source domains.
\begin{figure}
\centering
\subfigure[SSL]{
\includegraphics[width=3cm]{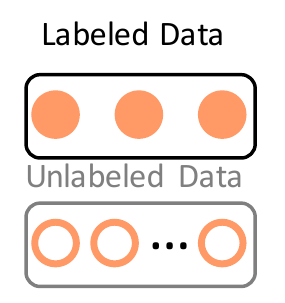}
}\hspace{12mm}
\subfigure[SSDG]{
\includegraphics[width=5cm]{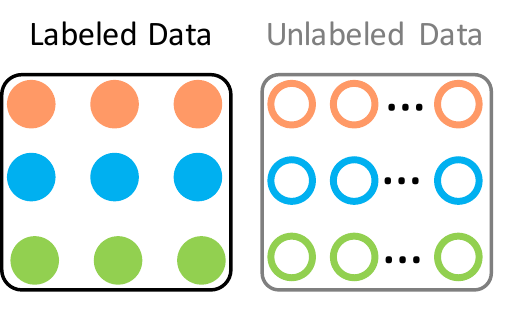}
}
\caption{Comparison between the typical semi-supervised learning (SSL) and the semi-supervised domain generalization (SSDG). Note that different colors denote different domains. In the SSDG setting, there are multiple training domains with different data distributions when compared with SSL.}
\label{fig04}
\end{figure}

Several domain generalization methods have been developed to handle this issue~\cite{DBLP:conf/cvpr/NamLPYY21,DBLP:conf/eccv/SeoSKKHH20,DBLP:conf/iccv/YueZZSKG19,wang2022feature,qi2022novel,DBLP:journals/pr/ZhangQSG22}. However, these methods need to label all data from source domains, which is expensive and time-consuming in real-world applications. In general, a few labels for a dataset are easy to obtain, thus the semi-supervised domain generalization (SSDG) task is recently proposed~\cite{DBLP:journals/corr/abs-2106-00592}, where there are a few labeled data in each source domain while most of the samples do not have the label information. To tackle this task, StyleMatch~\cite{DBLP:journals/corr/abs-2106-00592} is proposed via extending the FixMatch~\cite{DBLP:conf/nips/SohnBCZZRCKL20} with a couple of new ingredients, which resorts to enhancing the diversity from the image level and the classifier level. However, this method does not handle the data-distribution discrepancy of different source domains, resulting in a negative impact on the accuracy of pseudo-labels during training.

Similarly, in the conventional semi-supervised learning (SSL), most existing methods assume that all training samples are sampled from the same data distribution~\cite{DBLP:conf/nips/GrandvaletB04,DBLP:conf/iclr/LaineA17,DBLP:conf/nips/TarvainenV17,Nassar_2021_CVPR,Zheng_2022_CVPR,Wang_2022_CVPR,DBLP:conf/nips/ZhangWHWWOS21}. Differently, in the semi-supervised domain generalization task, the training samples from different distributions, as illustrated in Fig.~\ref{fig04}. Therefore, if directly using these methods to deal with the SSDG task, we cannot obtain the accurate pseudo-labels during the training course because the domain shift in the training set is not beneficial to producing the accurate pseudo-label.

In this paper, we mainly resort to obtaining accurate pseudo-labels so as to enhance the model's discrimination and generalization in the unseen domain. We first analyze the generalization error on a domain using the theory of multi-domain learning~\cite{ben2010theory}. Based on the upper bound of the generalization error, we can obtain a conclusion that: 1) alleviating the interference of different domains and 2) using all samples to train the model can effectively reduce the upper bound of generalization error on each domain, which means that the accurate pseudo-labels can be generated.

Inspired by the theory of multi-domain learning, we extend the FixMatch~\cite{DBLP:conf/nips/SohnBCZZRCKL20}~\footnote{FixMatch is an excellent baseline in SSDG, which will be validated in the experimental section.} to a multi-task learning method, named MultiMatch, for semi-supervised domain generalization. Specifically, we build the independent local task for each domain to mitigate the interference from different domains during generating pseudo-labels. In addition, we also construct a jointly global task for predicting unseen target domains. Particularly, each independent local task is utilized to give the pseudo-label, while the jointly global task is trained using the pseudo-label. Furthermore, benefiting from the multi-task framework, we can fuse the predictions from the global and local tasks to further improve the accuracy of pseudo-labels and the generalization capability of the model in the training and test stages, respectively. We conduct a series of experiments on several DG benchmark datasets. Experimental results demonstrate the effectiveness of the proposed method, and the proposed method can produce a performance improvement when compared with the semi-supervised learning methods and the semi-supervised DG method. Moreover, we also verify the efficacy of each module in our method.
In conclusion, our main contributions can be summarized as:
  \begin{itemize}
     \item We analyze the semi-supervised domain generalization task based on the generalization error of multi-domain learning, which inspires us to design a method to obtain the high-quality pseudo-label during training. 

    \item We propose a simple yet effective multi-task learning (\ie, MultiMatch) for semi-supervised domain generalization, which can effectively reduce the interference from different domains during pseudo-labeling. Also, most of the modules in the model are shared for all domains, which can be sufficiently trained by all samples. 
    To further promote the accuracy of pseudo-labels and the capability of the model, we propose to merge the outputs of the local task and the global task together to yield a robust prediction in the training and test stages.
    \item We evaluate our approach on multiple standard benchmark datasets, and the results show that our approach outperforms the state-of-the-art accuracy. Moreover, the ablation study and further analysis are provided to validate the efficacy of our method.
  \end{itemize}

The rest of this paper is organized as follows.
We review some related work in Section \ref{s-related}.
The proposed method is introduced in Section \ref{s-framework}.
Experimental results and analysis are presented in Section \ref{s-experiment},
and Section \ref{s-conclusion} is the conclusion.

\section{Related work}\label{s-related}
In this section, we investigate the related work to our work, including the semi-supervised learning methods and the domain generalization methods. The detailed investigation is presented in the following part.

\subsection{Semi-supervised Learning}
The semi-supervise learning has achieved remarkable performance in the recent years~\cite{DBLP:conf/nips/GrandvaletB04,DBLP:conf/iclr/LaineA17,DBLP:conf/nips/TarvainenV17,DBLP:conf/nips/BerthelotCGPOR19,DBLP:conf/iclr/BerthelotCCKSZR20,DBLP:conf/nips/SohnBCZZRCKL20,Li_2021_ICCV_Comatch,fu2021dynamic,Gong_2021_CVPR,Nassar_2021_CVPR,Zheng_2022_CVPR,Wang_2022_CVPR,DBLP:conf/nips/ZhangWHWWOS21,Yang_2022_CVPR,zhao2022balanced,wang2021pointwise}. For example,
Grandvalet \etal~\cite{DBLP:conf/nips/GrandvaletB04} propose to minimize entropy regularization, which enables to incorporate unlabeled data in the standard supervised learning.
Laine \etal~\cite{DBLP:conf/iclr/LaineA17} develop Temporal Ensembling, which maintains an exponential moving average of label predictions on each training example and penalizes predictions that are inconsistent with this target.
However, because the targets change only once per epoch, Temporal Ensembling becomes unwieldy when learning large datasets. To overcome this issue, Tarvainen \etal \cite{DBLP:conf/nips/TarvainenV17} design Mean Teacher that averages model weights instead of label predictions. In FixMatch \cite{DBLP:conf/nips/SohnBCZZRCKL20}, the method first generates pseudo-labels using the model’s predictions on weakly-augmented unlabeled images, and is then trained to predict the pseudo-label when fed a strongly-augmented version of the same image. Furthermore, Zhang \etal \cite{DBLP:conf/nips/ZhangWHWWOS21} introduce a curriculum learning approach to leverage unlabeled data according to the model’s learning status. The core of the method is to flexibly adjust thresholds for different classes at each time step to let pass informative unlabeled data and their pseudo-labels.

However, the success of the typical SSL largely depends on the assumption that the labeled and unlabeled data share an identical class distribution, which is hard to meet in the real-world application. The distribution mismatch between the labeled and unlabeled sets can cause severe bias in the pseudo-labels of SSL, resulting in significant performance degradation. To bridge this gap, Zhao \etal \cite{Zhao_2022_CVPR} put forward a new SSL learning framework, named Distribution Consistency SSL, which rectifies the pseudo-labels from a distribution perspective. 
Differently, Oh \etal \cite{Oh_2022_CVPR} propose a general pseudo-labeling framework that class-adaptively blends the semantic pseudo-label from a similarity-based classifier to the linear one from the linear classifier.

Different from these existing semi-supervised learning settings, we aim to address the semi-supervised task with multiple domains in the training procedure, which results in the interference of different domains during pseudo-labeling because of the impact of domain shift.

\subsection{Domain Generalization}
Recently, some methods are also developed to address the domain generalization problem in some computer vision tasks~\cite{DBLP:conf/cvpr/NamLPYY21,DBLP:conf/eccv/SeoSKKHH20,DBLP:conf/iccv/YueZZSKG19,DBLP:conf/cvpr/CarlucciDBCT19,DBLP:journals/tomccap/QiWHSG21,DBLP:conf/nips/BalajiSC18,DBLP:conf/iccv/LiZYLSH19,DBLP:conf/eccv/LiTGLLZT18,DBLP:journals/pr/ZhangQSG22,liu2021dg,Zhou_2022_CVPR}, such as classification and semantic segmentation.
Inspired by domain adaptation methods~\cite{DBLP:journals/tomccap/XuSDWXH22,wu2023domain}, some works based on domain alignment~\cite{DBLP:conf/eccv/LiTGLLZT18,DBLP:conf/nips/ZhaoGLFT20,DBLP:conf/icml/MuandetBS13,DBLP:journals/pr/RahmanFBS20,DBLP:conf/cvpr/LiPWK18,DBLP:conf/cvpr/GongLCG19,DBLP:conf/wacv/RahmanFBS19} aim at mapping all samples from different domains into the same subspace to alleviate the difference of data distribution across different domains. For example, Muandet~\etal~\cite{DBLP:conf/icml/MuandetBS13} introduce a kernel-based optimization algorithm to learn the domain-invariant feature representation and enhance the discrimination capability of the feature representation. However, this method cannot guarantee the consistency of the conditional distribution, hence Zhao~\etal~\cite{DBLP:conf/nips/ZhaoGLFT20} develop an entropy regularization term to measure the dependency between the learned feature and the class labels, which can effectively ensure the conditional invariance of learned feature, so that the classifier can also correctly classify the feature from different domains. 

 Besides, Gong~\etal~\cite{DBLP:conf/cvpr/GongLCG19} exploit CycleGAN~\cite{DBLP:conf/iccv/ZhuPIE17} to yield new styles of images that cannot be seen in the training set, which smoothly bridge the gap between source and target domains to boost the generalization of the model. Rahman~\etal~\cite{DBLP:conf/wacv/RahmanFBS19} also employ GAN to generate synthetic data and then mitigate domain discrepancy to achieve domain generalization. Differently, Li~\etal~\cite{DBLP:conf/cvpr/LiPWK18} adopt an adversarial auto-encoder learning framework to learn a generalized latent feature representation in the hidden layer, and use Maximum Mean Discrepancy to align source domains, then they match the aligned distribution to an arbitrary prior distribution via adversarial feature learning. In this way, it can better generalize the feature of the hidden layer to other unknown domains. Rahman~\etal~\cite{DBLP:journals/pr/RahmanFBS20} incorporate the correlation alignment module along with adversarial learning to help achieve a more domain-agnostic model because of the improved ability to more effectively reduce domain discrepancy.
 In addition to performing adversarial learning at the domain level to achieve domain alignment, Li~\etal~\cite{DBLP:conf/eccv/LiTGLLZT18} also conduct domain adversarial tasks at the 
class level to align samples of each category from different domains.

Ideally, visual learning methods should be generalizable, for dealing with any unseen domain shift when deployed in a new target scenario, and they should be data-efficient for reducing development costs by using as few labels as possible. 
However, the conventional DG methods, which are unable to handle unlabeled data, perform poorly with limited labels in the SSDG task. 
To handle this task, Zhou \etal \cite{DBLP:journals/corr/abs-2106-00592} propose StyleMatch, a simple approach that extends FixMatch with a couple of new ingredients tailored for SSDG: 1) stochastic modeling for reducing overfitting in scarce labels, and 2) multi-view consistency learning for enhancing the generalization capability of the model. However, StyleMatch cannot effectively address the interference of different domains during pseudo-labeling. In this paper, we deal with the SSDG task from the multi-task learning perspective.

\section{The proposed method}\label{s-framework}

\begin{figure*}[t]
\centering
\includegraphics[width=14cm]{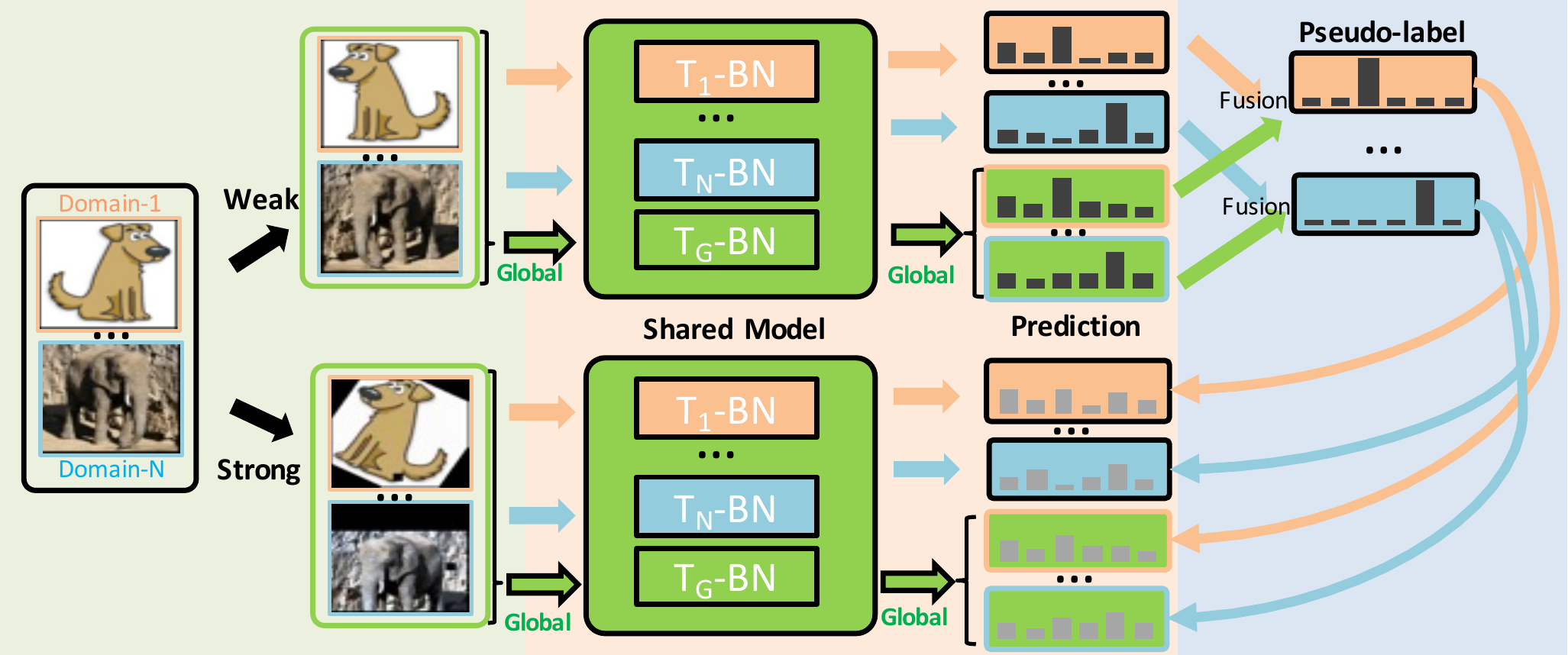}
\caption{An illustration of the proposed MultiMatch for the semi-supervised domain generalization setting. In the method, training each domain is considered as the local task, and training all domains together is viewed as the global task. In the training course, we propose to employ the prediction fusion scheme to generate the final pseudo-label, which can be used to train the unlabeled samples. For example, we utilize the images based on weak augmentation to generate the pseudo-label. For an image from the $i$-th domain, we can obtain two predictions from the ``$\mathrm{T_i}$-BN'' and ``$\mathrm{T_G}$-BN'' paths (\ie, the local and global tasks), then we can fuse the two predictions to generate the final pseudo-label of the image by Function~\ref{f01} (\ie, choose the highest confidence as pseudo-labels). For the image of the strong  augmentation, we can obtain two predictions from the ``$\mathrm{T_i}$-BN'' and ``$\mathrm{T_G}$-BN'' paths, and use the pseudo-label of the image to compute the loss and update the network.}
\label{fig01}
\vspace{-12pt}
\end{figure*}

In this paper, our goal is to address the semi-supervised domain generalization (SSDG) task. In this task, a few labeled samples on each domain are available, while most of the samples lack the label information, thus a key task is how to generate the accurate pseudo-label. Different from the conventional semi-supervised methods, the SSDG task contains multiple source domains with different distributions. We first analyze the generalization error of multi-domain learning, which shows the upper bound of the generalization error is related to the discrepancy of different domains and the number of training samples. Particularly, reducing the upper bound is equal to improving the accuracy of the pseudo-label.  Based on the analysis, we develop a multi-task learning method (\ie, MultiMatch) to deal with the SSDG task, as illustrated in Fig.~\ref{fig01}. Besides, to further leverage the advantage of the multi-task learning framework, we fuse the predictions from different tasks to yield a better pseudo-label and more generalizable model in the training and test procedure. We will introduce our method in the following part.

\subsection{Theoretical Insight}
In the semi-supervised DG task, each domain contains labeled data and unlabeled data. To be specific, given $N$ domains in the training stage, we use $\mathcal{D}_i=\{\mathcal{D}_i^l, \mathcal{D}_i^u\}$ to denote the $i$-th domain, where $\mathcal{D}_i^l$ and $\mathcal{D}_i^u$ represent the labeled and unlabeled samples in the $i$-th domain, respectively. Since there is no label information for $\mathcal{D}_i^u$, we aim to generate the high-quality pseudo-label in the training stage. Particularly, the SSDG task in the training stage can also be considered as multi-domain learning, and these unlabeled data can be viewed as the test data during pseudo-labeling. In the next part, we will introduce a theory of multi-domain learning to explore the semi-supervised DG task from the theoretical perspective.

Here,  we consider hypotheses $h \in \mathcal{H}$ (\ie, prediction function), and give a vector $\alpha=(\alpha_1,\ldots,\alpha_N)$ of domain weights with $\sum_{j=1}^N \alpha_j=1$. In addition, we assume that the learner receives a total of $m$ labeled training samples, with $m_j=\beta_jm$ from the $j$-th domain $\mathcal{D}_j$. We define the empirical $\alpha$-weighted error of function $h$ as

\begin{equation}
  \begin{aligned}
\hat{\epsilon}_{\alpha}(h)=\sum_{j=1}^N\alpha_j\hat{\epsilon}_{j}(h)=\sum_{j=1}^N\frac{\alpha_j}{m_j}\sum_{x\in S_j}|h(x)-f_j(x)|, 
    \end{aligned}
  \label{eq03}
  \end{equation}
  where $f_j(x)$ is a labeling function for the $j$-th domain (\ie, the mapping from a sample to its ground truth).

\textbf{Theorem 1}~\cite{ben2010theory} Let $\mathcal{H}$ be a hypothesis space of VC dimension~\cite{hastie2009elements} $d$. For each domain $j \in \{1,...,N\}$,
let $S_j$ be a labeled sample of size $\beta_j m$ generated by drawing $\beta_j m$ points from $\mathcal{D}_j$ and
labeling them according to $f_j(\cdot)$. If $\hat{h} \in \mathcal{H}$ is the empirical minimizer of $\hat{\epsilon}_{\alpha}(h)$ for a fixed
weight vector $\alpha$ on these samples and $h_T^*=\min_{h\in\mathcal{H}}\epsilon_T(h)$ is the target error minimizer, then
the upper bound of the generalization error on the target domain $\mathcal{D}_T$, for any $\delta \in (0, 1)$, with probability at least $1- \delta$, can be written as

\begin{equation}
  \begin{aligned}
\epsilon_{T}(\hat{h})\leqslant \epsilon_{T}(h_T^*)+2\sqrt{(\sum_{j=1}^{N}\frac{\alpha_j^2}{\beta_j})\frac{d\log(2m)+\log(\delta)}{2m}}\\
+\sum_{j=1}^N\alpha_j(2\lambda_j+d_{\mathcal{H}\Delta\mathcal{H}}(\mathcal{D}_j, \mathcal{D}_T)),
    \end{aligned}
  \label{eq01}
  \end{equation}
where 
\begin{equation}
  \begin{aligned}
\lambda_j=\min_{h\in \mathcal{H}}\{\epsilon_{T}(h)+\epsilon_{j}(h)\}. 
    \end{aligned}
  \label{eq02}
  \end{equation}
The detailed demonstration process can be found in ~\cite{ben2010theory}.

 In Eq.~\ref{eq01}, the third term indicates the distribution discrepancy between $\mathcal{D}_j$ and $\mathcal{D}_T$. It is worth noting that, in the semi-supervised DG setting, all $\{\alpha_i\}_{i=1}^{N}$ and $\{\beta_i\}_{i=1}^{N}$ are $\frac{1}{N}$. Therefore, the Eq.~\ref{eq01} can be rewritten as

 \begin{equation}
  \begin{aligned}
\epsilon_{T}(\hat{h})\leqslant \epsilon_{T}(h_T^*)+\sum_{j=1}^N\frac{2\lambda_j}{N}+2\sqrt{\frac{d\log(2m)+\log(\delta)}{2m}}\\
+\sum_{j=1}^N\frac{1}{N}d_{\mathcal{H}\Delta\mathcal{H}}(\mathcal{D}_j, \mathcal{D}_T).
    \end{aligned}
  \label{eq04}
  \end{equation}

According to Eq.~\ref{eq04}, the upper bound of the generalization error on the target domain is mainly decided by four terms, where the first two terms can be considered as the constant. There are two observations in the third term and the fourth term. 1) In the third term, when $m$ is a larger value, the total value is smaller, which indicates that we should use the available samples to train the model. In other words, most modules in the designed model should be shared for all domains or tasks. 2) The last term is the distribution discrepancy between the source-target domain~\footnote{When the model generates the pseudo-label for a source domain, this source domain is named the source-target domain.} and the other domains. If we can alleviate the discrepancy, the generalization error on the source-target domain can also be reduced. 

Particularly, since the semi-supervised DG task consists of multiple domains, it can be also viewed as the multi-domain learning paradigm in the training stage. Besides, the unlabeled training data can be also considered as the test data during pseudo-labeling. Therefore, it means that, if we can effectively reduce the upper bound of the generalization error in Eq.~\ref{eq04}, it will result in generating a high-quality pseudo-label for each domain in the training stage.

\subsection{Multi-task Learning Framework}
FixMatch is a significant simplification of existing SSL methods~\cite{DBLP:conf/nips/SohnBCZZRCKL20}. It first generates pseudo-labels using the model’s predictions on weakly-augmented unlabeled images. For a given image, the pseudo-label is only retained if the model produces a high-confidence prediction. The model is then trained to predict the pseudo-label when fed a strongly-augmented version of the same image. Despite its simplicity, FixMatch achieves state-of-the-art
performance across a variety of standard semi-supervised learning benchmarks. In the SSDG task, FixMatch also achieves better performance than other conventional semi-supervised methods~\cite{DBLP:conf/nips/ZhangWHWWOS21,Wang_2022_CVPR}, which is verified in our experiment. In this paper, we aim to extend FixMatch to a multi-task framework.

Based on the upper bound of the generalization error in Eq.~\ref{eq04}, the proposed method needs to satisfy two requirements: 1) most of modules in the model are shared for all domains, which can be sufficiently trained by all samples, and 2) the model can reduce the interference of domain gap between different domains. Therefore, we propose a multi-task learning framework, named MultiMatch, to address the SSDG task.

In this part, we will describe our multi-task learning framework. Since there are multiple different domains in the SSDG task, we consider training each domain as an independent task (\ie, the local task for each domain), which can effectively reduce the interference between different domains during pseudo-labeling. Besides, considering the SSDG task also needs to exploit the trained model on the unseen domain, we add a global task where we employ pseudo-labels from the local task to train the model. We assume that there are $N$ source domains in the training stage, thus we will build $N+1$ tasks in our framework. To be specific, we will employ the independent batch normalization (BN)~\cite{DBLP:conf/cvpr/ChangYSKH19} and classifier for each task in our method, and other modules are shared for all tasks.
The batch normalization (BN) can be formulated by
    \begin{equation}
  \begin{aligned}
  &{\rm BN}(f_d)= \gamma \frac{f_d-M_\mu(\phi)}{M_{\sigma}(\phi)}+\beta,~~d \in \phi,
  \end{aligned}
    \label{eq05}
  \end{equation}
  where $\phi$ is the domain set, and $M_{\mu}(\phi)$ and $M_{\sigma}(\phi)$ represent the statistics for the $\phi$. When $\phi$ merely includes a domain, the sample will be normalized by the statistics from the own domain. Differently, when $\phi$ includes all domains, each sample will be normalized by the shared statistics from all domains. Since there exists a domain gap across different domains, the shared statistics will bring the noise error, as shown in the last term of Eq.~\ref{eq04}, resulting in larger generalization error.

\textit{Remark.} In our multi-task learning framework, using the independent BN can effectively mitigate the interference of different domains, as shown in Eq.~\ref{eq04}. In addition, in our method, most of the modules are shared for all domains, which can sufficiently exploit all samples to reduce the third item in Eq.~\ref{eq04}. Hence, our multi-task learning framework can obtain a small generalization error on each domain so as to generate an accurate pseudo-label for each domain in the training stage. Last but not least, the multi-task framework with multiple classifiers can produce the ensemble prediction for pseudo-label and test evaluation in the training and test procedure. We will give the prediction fusion scheme in the next part.

\subsection{Prediction Fusion}\label{sec:PF}
For an image $x_i$ from the $i$-th domain, we can generate its output $Y_i=[y_1; \ldots; y_{N+1}] \in \mathbb{R}^{C\times (N+1)}$, where $C$ is the number of the classes. In the training stage, due to the interference of different domains with each other, we employ the output of the $i$-th task as the main prediction. To further guarantee the reliability of pseudo-labels in the training course, we combine it with the output of the global task to generate the final pseudo-labels. The motivation of fusing the the global prediction is that the local and global predictions can be viewed as the private and shared information for all images, thus they are complementary. In this paper, we fuse them together with the Function~\ref{f01} (\ie, choose the highest confidence as pseudo-labels) to get the accurate pseudo-label. For an image $x_i$ from the $i$-th domain, the pseudo-label is produced by

    \begin{equation}
  \begin{aligned}
  \hat{y} = {\rm PredictLabel} ([y_i; y_{N+1}]),
  \end{aligned}
    \label{eq06}
  \end{equation}
  where ${\rm PredictLabel}(\cdot)$ is defined in Function~\ref{f01}. In the test stage, we use the output of the global task as the main perdition because the test domain is unseen (\ie, we do not know which training domain is similar to the test domain). Besides, for a test image, we also fuse the output of the most similar task to yield the final prediction result as

   \begin{equation}
  \begin{aligned}
  y_{max} = {\rm SelectTask} ([y_1; \ldots; y_N]);
  \end{aligned}
    \label{eq07}
  \end{equation}

     \begin{equation}
  \begin{aligned}
   \tilde{y}  = {\rm PredictLabel} (\frac{y_{max}+y_{N+1}}{2}),
  \end{aligned}
    \label{eq08}
  \end{equation}
  where ${\rm SelectTask}(\cdot)$ is defined in Function~\ref{f02} and returns the prediction of the most similar task (\ie, the column has the maximum value in a matrix) for a test image in the test procedure.

\floatname{algorithm}{Function}
\begin{algorithm}
\caption{\small{PredictLabel}}~\label{alg01}
\label{f01}
\begin{algorithmic}[1]
\STATE {\bf Input:} A matrix $Y\in \mathbb{R}^{c\times n}$.\\
\STATE {\bf Output:} A vector $y\in \mathbb{R}^{c\times 1}$. \\
\STATE Find the maximum value of each line in $Y$ to generate a vector $y\in \mathbb{R}^{c\times 1}$.\\
\STATE Find the position (\eg, $p$) of the maximum value of the vector $y$.\\
\STATE Setting the position ($p$) of $y$ as 1 (\ie, y(p)=1), and other positions as 0.\\
\STATE Return $y$.
\end{algorithmic}
\end{algorithm}

\begin{algorithm}
\caption{\small{SelectTask}}~\label{alg01}
\label{f02}
\begin{algorithmic}[1]
\STATE {\bf Input:} A matrix $Y\in \mathbb{R}^{c\times n}$.\\
\STATE {\bf Output:} A vector $y\in \mathbb{R}^{c\times 1}$. \\
\STATE Find the column's position (\eg, $p$) of the maximum value of the vector $Y$.
\STATE Select the $p$-th column from $Y$ as $y$ (\ie, $y= Y(:, p)$).
\STATE Return $y$.
\end{algorithmic}
\end{algorithm}

\subsection{Training Process}
In our method, we merely use the cross-entropy loss to train our model as in FixMatch~\cite{DBLP:conf/nips/SohnBCZZRCKL20}. In the training course, we randomly select the same number of labeled and unlabeled images from each domain to form a batch. It is worth noting that, each image passes the domain-specific BN and classifier, and all images are required to pass the global task. Simultaneously, we fuse the predictions of the domain-specific task and the global task together to generate accurate pseudo-labels. The overall training process is shown in Algorithm~\ref{alg01}.

\floatname{algorithm}{Algorithm}
\begin{algorithm}[ht]
\setcounter{algorithm}{0}
\caption{\small{The training procedure of the proposed MultiMatch}}~\label{alg01}
\begin{algorithmic}[1]
\STATE {\bf Input:} All training domains $\{\mathcal{D}_i^l, \mathcal{D}_i^u\}_{i=1}^N$.\\
\STATE {\bf Output:} The trained parameters of the model $\theta$. \\
\STATE {\bf Initialization:} Initialize the parameters $\theta$. \\
\FOR{iter $=<$ MaxIter}
\STATE Generating a batch of images $\{\mathcal{B}^l, \mathcal{B}^u\}.$
\STATE Using the labeled samples $\mathcal{B}^l$ to train the model, each image passes the domain-specific BN and classifier, and all images pass the global BN and classifier.
\STATE Conducting weak and strong augmentations for $\mathcal{B}^u$ to generate $\mathcal{B}_w^u$ and $\mathcal{B}_s^u$.
\STATE Based on $\mathcal{B}_w^u$ to yield the pseudo-label $Y_u$ by Eq.~\ref{eq06}.
\STATE Using $\mathcal{B}_s^u$ and $Y_u$ to train the model, and the forward path of all images is consistent with $\mathcal{B}^l$.
\ENDFOR
\end{algorithmic}
\label{al01}
\end{algorithm}

\section{Experiments}\label{s-experiment}

In this section, we first introduce the experimental datasets and settings in Section~\ref{sec:EXP-DS}. Then, we compare the proposed MultiMatch with the state-of-the-art SSL methods and the SSDG method in Section~\ref{sec:EXP-CUA}, respectively. To validate the effectiveness of various components in our MultiMatch, we conduct ablation studies in Section~\ref{sec:EXP-SS}. Lastly, we further analyze the property of the proposed method in Section~\ref{sec:EXP-FA}.
\subsection{Datasets and Experimental Settings}\label{sec:EXP-DS}
\subsubsection{Datasets} 
\begin{figure}
\centering
\subfigure[PACS]{
\includegraphics[width=5.51cm]{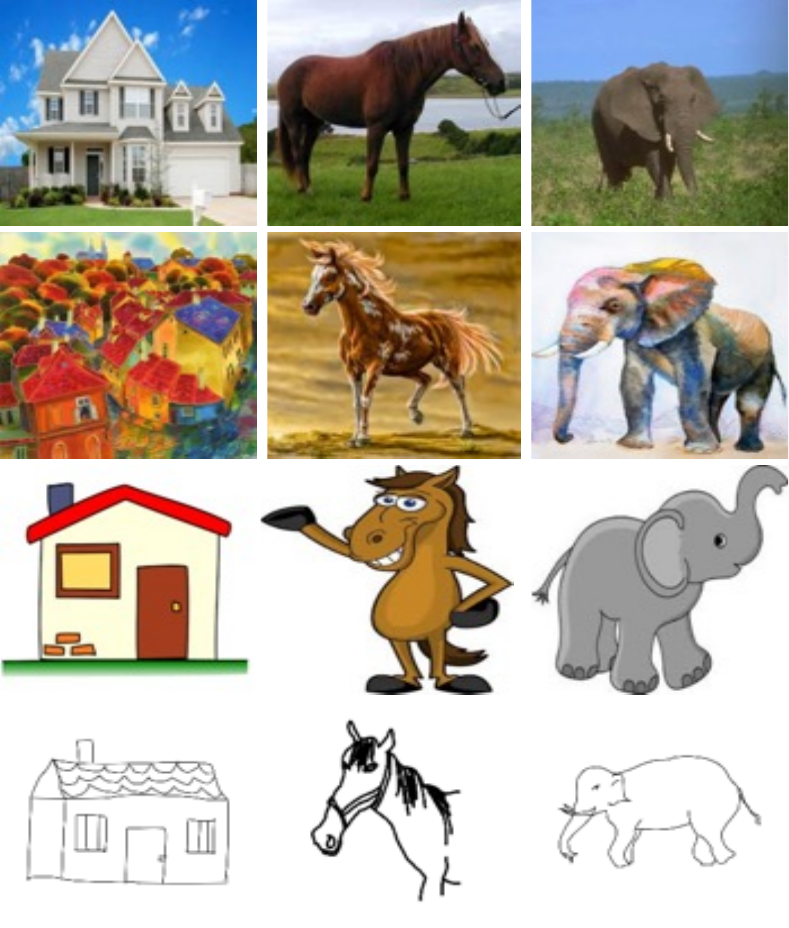}
}\hspace{10mm}
\subfigure[Office-Home]{
\includegraphics[width=5.51cm]{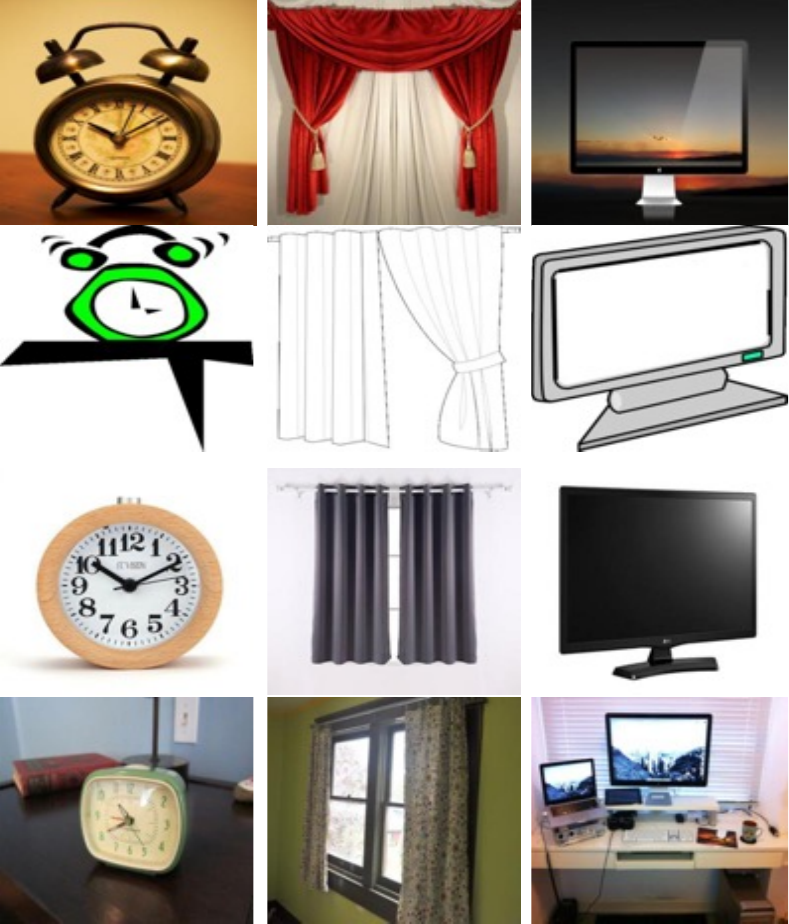}
}
\caption{Examples on PACS and Office-Home, respectively. It is worth noting that different rows denote different domains.}
\label{fig05}
\end{figure}
\begin{figure}
\centering
\subfigure[Horse]{
\includegraphics[width=5.51cm]{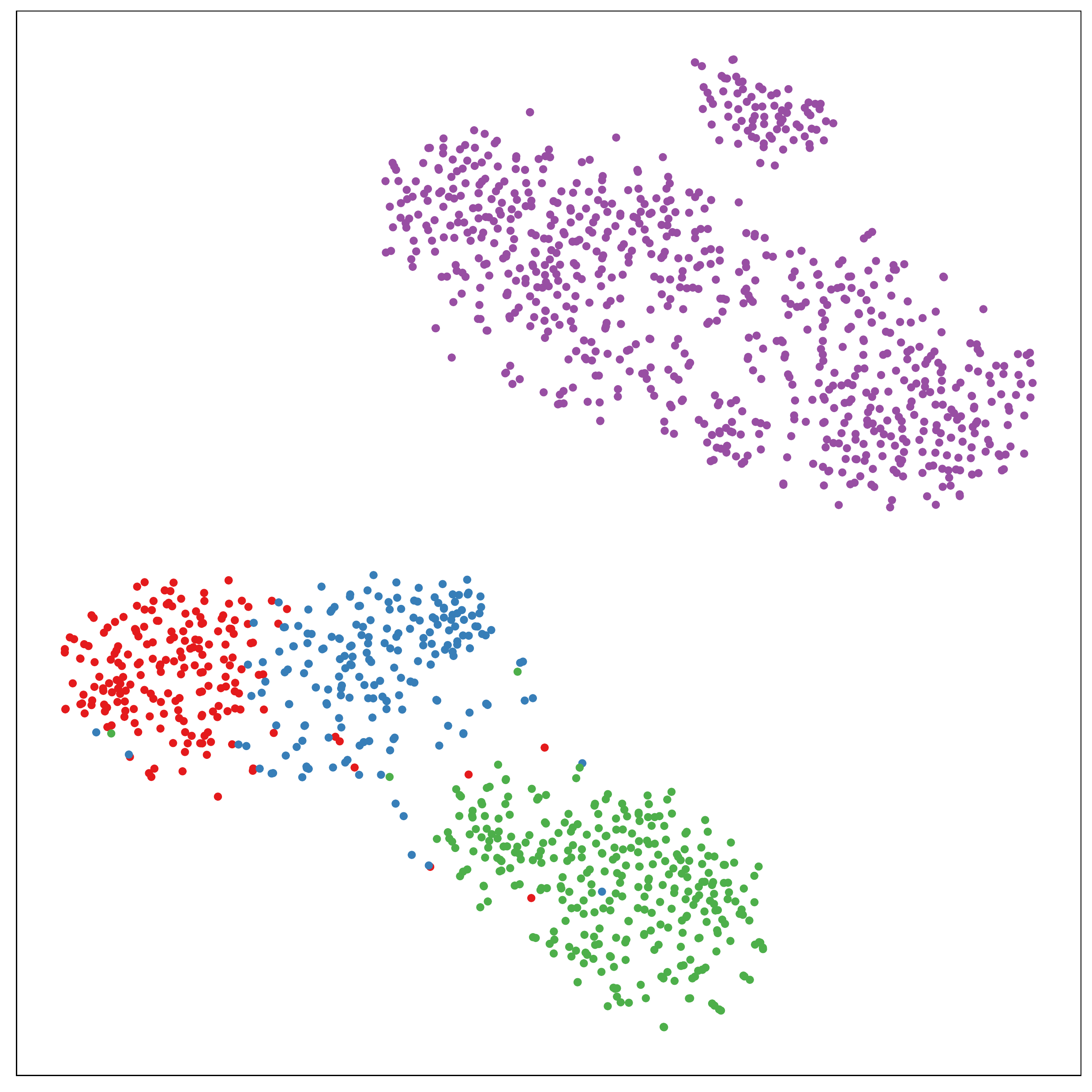}
}
\subfigure[Dog]{
\includegraphics[width=5.51cm]{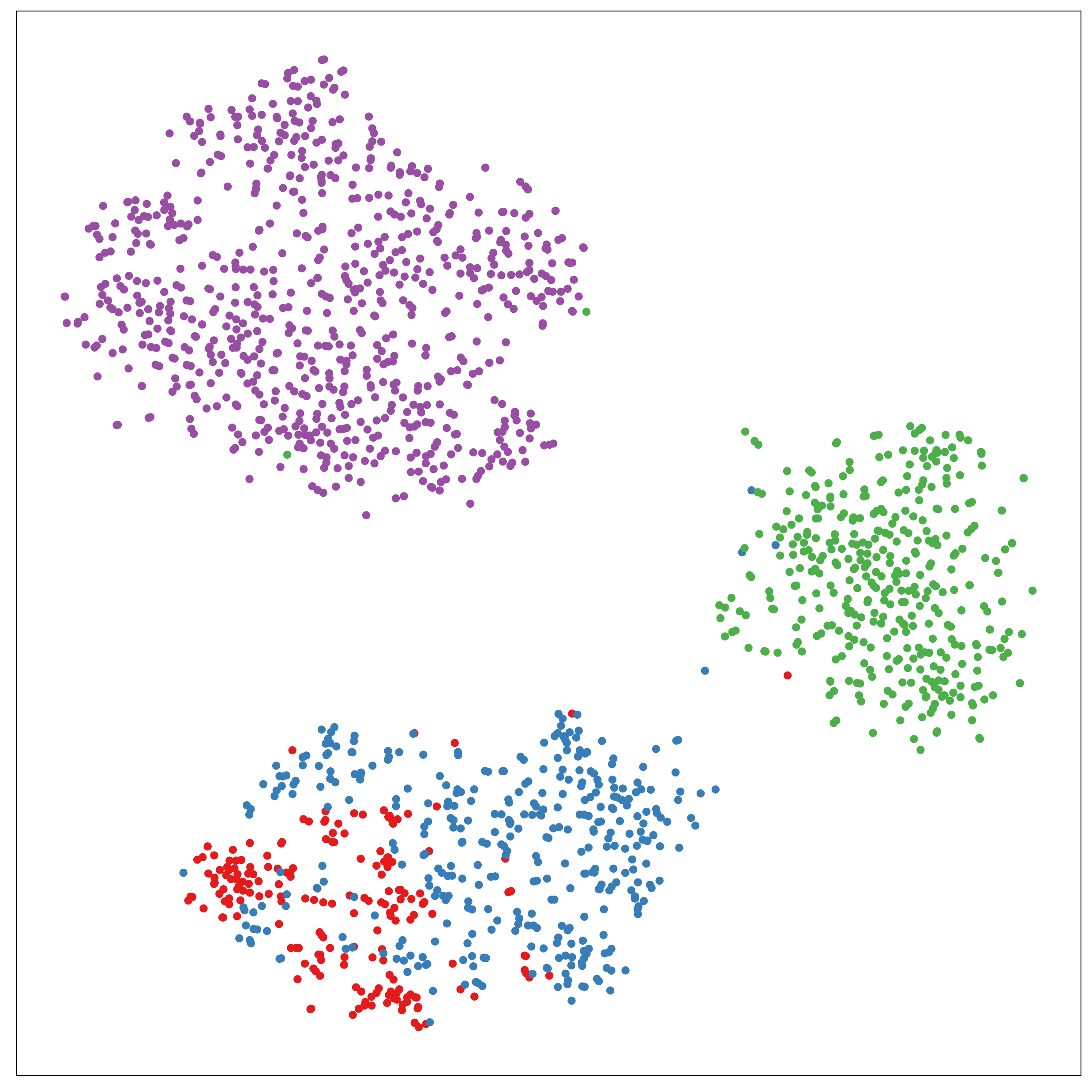}
}
\subfigure[Elephant]{
\includegraphics[width=5.51cm]{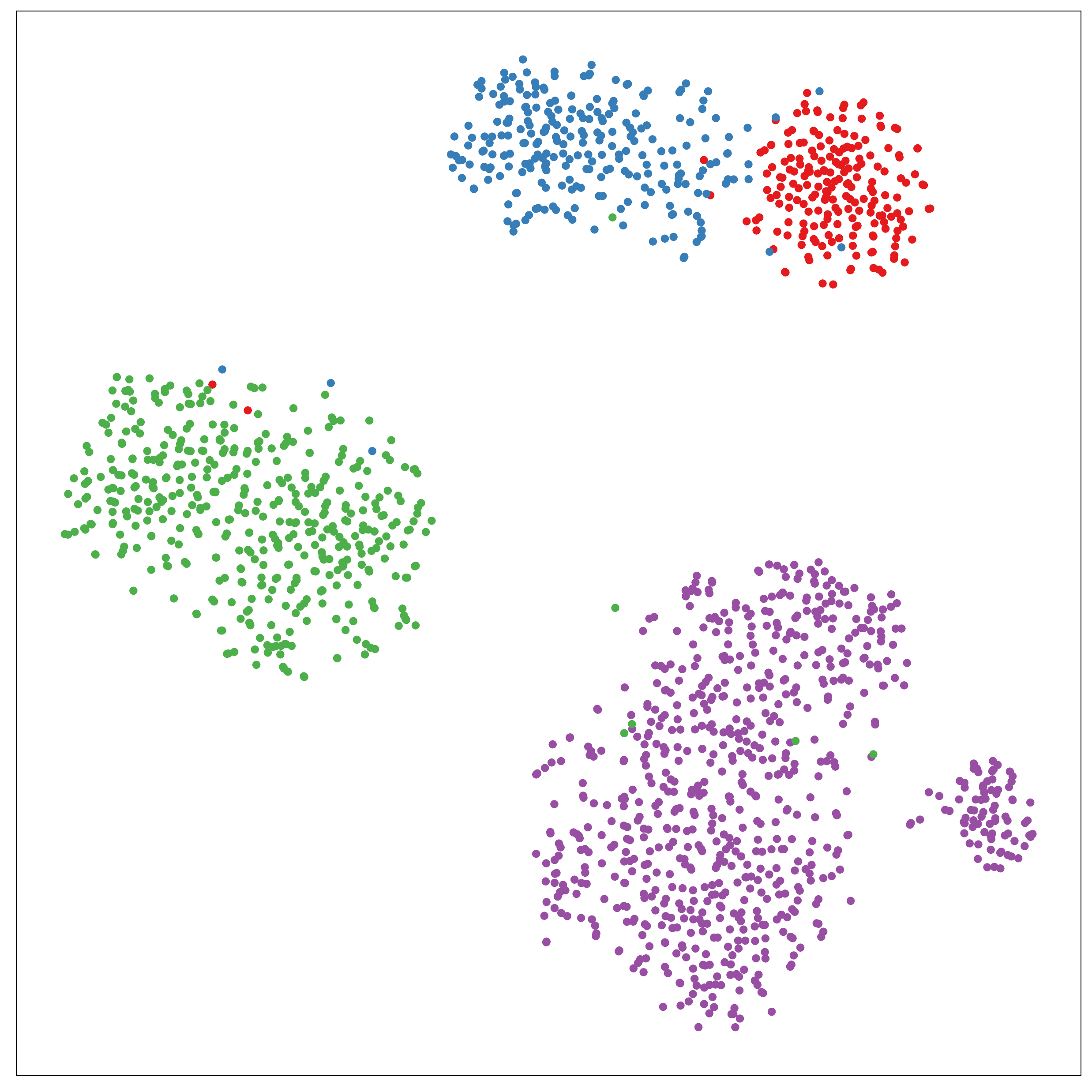}
}
\subfigure[Giraffe]{
\includegraphics[width=5.51cm]{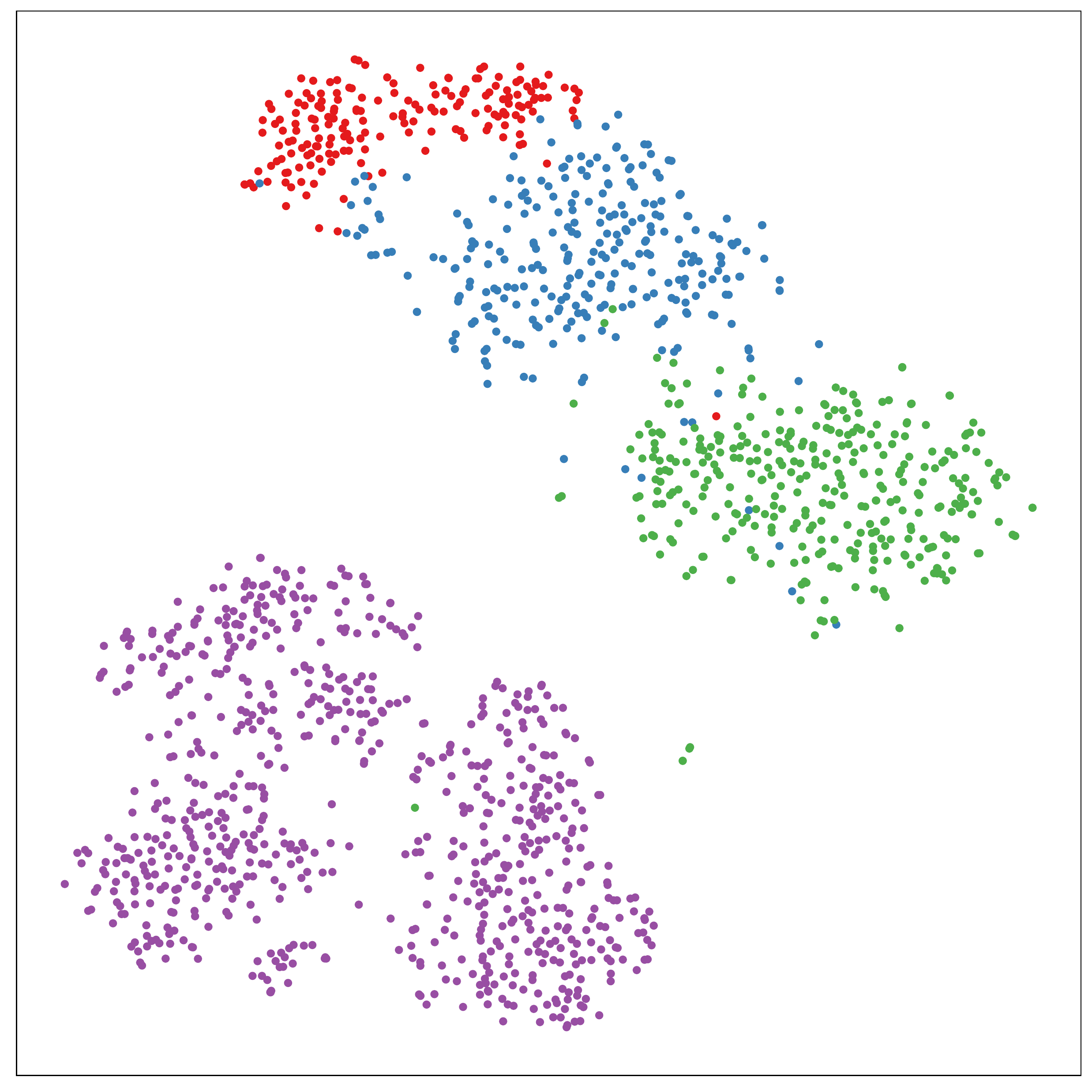}
}
\caption{Visualization of the feature representation on four classes of PACS by t-SNE~\cite{van2008visualizing}. The features are extracted by ResNet18 pre-trained on ImageNet~\cite{DBLP:conf/cvpr/DengDSLL009}. Note that different colors indicate different domains.}
\label{fig07}
\end{figure}

In this paper, we conduct the experiments to validate the effectiveness of our method on three benchmark DG datasets as follows:
  \begin{itemize}
      \item \textbf{PACS} \cite{Li2017DeeperBA} consists of four different domains: Photo, Art, Cartoon and Sketch. It contains 9,991 images with 7 object categories in total, including Photo (1,670 images), Art (2,048 images), Cartoon (2,344 images), and Sketch (3,929 images).
      \item \textbf{Office-Home} \cite{venkateswara2017deep} contains 15,588 images of 65 categories of office and home objects. It has four different domains namely Art (2,427 images), Clipart (4,365 images), Product (4,439 images) and Real World (4,357 images), which is originally introduced for UDA but is also applicable in the DG setting.
      \item \textbf{miniDomainNet} \cite{DBLP:journals/tip/ZhouYQX21} takes a subset of DomainNet~\cite{DBLP:conf/iccv/PengBXHSW19} and utilizes a
smaller image size (96 $\times$ 96). miniDomainNet includes four domains and 126 classes. As a result, miniDomainNet contains 18,703 images of Clipart, 31,202 images of Painting, 65,609 images of Real and 24,492 images of Sketch.
  \end{itemize}
We show some examples from PACS and Office-home in Fig.~\ref{fig05}. As seen, there is an obvious difference among different domains. Besides, we also visualize the features of four categories on PACS by t-SNE~\cite{van2008visualizing}, as illustrated in Fig.~\ref{fig07}. In this figure, different colors are different domains. We observe that different domains appear in different spaces, which validates that there exists a domain shift in the training set.

\subsubsection{Implementation Details} 
Following the common practice~\cite{DBLP:conf/eccv/HuangWXH20}, ResNet18~\cite{DBLP:conf/cvpr/HeZRS16} pre-trained on ImageNet~\cite{DBLP:conf/cvpr/DengDSLL009} is employed as the CNN backbone (for all models compared in this paper). We randomly sample 64 images from each source domain to construct a minibatch for labeled and unlabeled data, respectively. The initial learning rate is set as $0.003$ for the pre-trained backbone. In the experiment, to enrich the diversity of the augmentation, we integrate the AdaIN augmentation~\cite{DBLP:conf/iccv/HuangB17} into the strong augmentation scheme. Particularly, the augmentation scheme is utilized in all experiments, including the baseline (\ie, FixMatch), thus it is a fair comparison in the experiment. 

\subsection{Comparison with State-of-the-art Methods}\label{sec:EXP-CUA}
We conduct the experiment to compare our method with some semi-supervised learning methods (\ie,  MeanTeacher~\cite{DBLP:conf/nips/TarvainenV17}, EntMin~\cite{DBLP:conf/nips/GrandvaletB04}, DebiasPL~\cite{Wang_2022_CVPR}, FlexMatch~\cite{DBLP:conf/nips/ZhangWHWWOS21}, FixMatch~\cite{DBLP:conf/nips/SohnBCZZRCKL20}) and the semi-supervised DG method (\ie, StyleMatch~\cite{DBLP:journals/corr/abs-2106-00592}). In this experiment, we run these methods under two different settings (\ie, 10 labels per class and 5 labels per class) on three benchmark datasets. The experimental results are reported in Tables~\ref{tab01}and~\ref{tab11}. As seen in this table, among these typical semi-supervised methods, FixMatch can achieve the best performance. Compared with FixMatch, our method can further improve performance. For example, on PACS our method outperforms FixMatch by $+3.74\%$ ($81.57$ vs. $77.83$) and $+5.09\%$ ($80.04$ vs. $74.95$) under the ``10 labels per class'' case and the ``5 labels per class'' case, respectively. Besides, on the large-scale dataset (\ie, miniDomainNet), our method can also achieve an obvious improvement, which is attributed to the fact that our method reduces the interference of different domains while guaranteeing that all training samples are utilized to train the model.

In addition, StyleMatch is developed to address the semi-supervised domain generalization task. Compared with it, our method has better experimental results on all settings and datasets, except for the ``5 labels per class'' case on PACS. For example, on miniDomainNet, our method increases StyleMatch by $+3.66\%$ ($58.79$ vs. $55.13$) and $+3.57\%$ ($54.61$ vs. $51.04$) under the ``10 labels per class'' case and the ``5 labels per class'' case, respectively. This confirms the effectiveness of our method when compared with the SOTA method, thanking the advantage of the multi-task learning framework in the semi-supervised domain generalization task. The effectiveness of each module in our proposed MultiMatch will be verified in the next ablation study. 

\begin{table*}[htbp]
  \centering
  \caption{Comparison between our method and different semi-supervised (DG) methods under different numbers of labeled samples on PACA and Office-Home. Note that ``P'', ``A'', ``C'' and ``S'' denote different domains on PACS. ``Avg'' is the average result of all domains. The \textbf{bold} is the best result.}
    \begin{tabular}{l|C{0.65cm}C{0.65cm}C{0.65cm}C{0.65cm}C{0.65cm}|C{0.65cm}C{0.65cm}C{0.65cm}C{0.65cm}C{0.65cm}}
    \toprule
          & \multicolumn{5}{c|}{PACS}             & \multicolumn{5}{c}{Office-Home} \\
    \midrule
    \multicolumn{1}{c|}{\multirow{2}[4]{*}{Method}} & P     & A     & C     & S     & Avg   & A     & C     & P     & R     & Avg \\
\cmidrule{2-11}          & \multicolumn{10}{c}{10 labels per class} \\
    \midrule
    MeanTeacher~\cite{DBLP:conf/nips/TarvainenV17} & 85.95 & 62.41 & 57.94 & 47.66 & 63.49 & 49.92 & 43.42 & 64.61 & 68.79 & 56.69  \\
    EntMin~\cite{DBLP:conf/nips/GrandvaletB04} & 89.39 & 72.77 & 70.55 & 54.38 & 71.77 & 51.92 & 44.92 & 66.85 & 70.52 & 58.55  \\
    FlexMatch~\cite{DBLP:conf/nips/ZhangWHWWOS21} & 66.04 & 48.44 & 60.79 & 53.23 & 57.13 & 25.63 & 28.25 & 37.98 & 36.15 & 32.00 \\
    DebiasPL~\cite{Wang_2022_CVPR} & 93.23 & 74.60 & 67.23 & 63.49 & 74.64 & 44.71 & 38.62 & 64.42 & 68.39 & 54.04 \\
    FixMatch~\cite{DBLP:conf/nips/SohnBCZZRCKL20} & 87.40 & 76.85 & 69.40 & 77.68 & 77.83 & 49.90 & 50.98 & 63.79 & 66.75 & 57.86 \\
    StyleMatch~\cite{DBLP:journals/corr/abs-2106-00592} & 90.04 & 79.43 & \textbf{73.75} & \textbf{78.40} & 80.41 & 52.82 & 51.60 & 65.31 & 68.61 & 59.59\\
    MultiMatch (ours)  & \textbf{93.25} & \textbf{83.30} & 73.00 & 76.74 & \textbf{81.57} & \textbf{56.32} & \textbf{52.76} & \textbf{68.94} & \textbf{72.28} & \textbf{62.57} \\
    \midrule
          & \multicolumn{10}{c}{5 labels per class} \\
    \midrule
    MeanTeacher~\cite{DBLP:conf/nips/TarvainenV17} & 73.54 & 56.00 & 52.64 & 36.97 & 54.79 & 44.65 & 39.15 & 59.18 & 62.98 & 51.49 \\
    EntMin~\cite{DBLP:conf/nips/GrandvaletB04} & 79.99 & 67.01 & 65.67 & 47.96 & 65.16 & 48.11 & 41.72 & 62.41 & 63.97 & 54.05 \\
    FlexMatch~\cite{DBLP:conf/nips/ZhangWHWWOS21} & 61.80 & 23.83 & 37.28 & 48.09 & 42.75 & 19.04 & 26.16 & 30.81 & 23.00 & 24.75 \\
     DebiasPL~\cite{Wang_2022_CVPR} & 89.52 & 65.52 & 54.09 & 29.22 & 59.59 & 41.41 & 33.45 & 58.77 & 60.59 & 48.56 \\
    FixMatch~\cite{DBLP:conf/nips/SohnBCZZRCKL20} & 86.15 & 74.34 & 69.08 & 73.11 & 75.67 & 46.60 & 48.74 & 59.65 & 63.21 & 54.55 \\
    StyleMatch~\cite{DBLP:journals/corr/abs-2106-00592} & 89.25 & 78.54 & \textbf{74.44} & \textbf{79.06} & \textbf{80.32} & 51.53 & 50.00 & 60.88 & 64.47 & 56.72 \\
    MultiMatch (ours)   & \textbf{90.66} & \textbf{82.43} & 72.43 & 74.64 & 80.04 & \textbf{53.71} & \textbf{50.21} & \textbf{63.99} & \textbf{67.81} & \textbf{58.93} \\
    \bottomrule
    \end{tabular}%
  \label{tab01}%
  \vspace{-5pt}
\end{table*}%

\begin{table*}[htbp]
  \centering
  \caption{Comparison between our method and different semi-supervised (DG) methods under different numbers of labeled samples on miniDomainNet. The \textbf{bold} is the best result.}
    \begin{tabular}{l|ccccc}
    \toprule
          & \multicolumn{5}{c}{miniDomainNet} \\
    \midrule
    \multicolumn{1}{c|}{\multirow{2}[4]{*}{Method}} & C     & P     & R     & S     & Avg \\
\cmidrule{2-6}          & \multicolumn{5}{c}{10 labels per class} \\
    \midrule
    MeanTeacher~\cite{DBLP:conf/nips/TarvainenV17} & 46.75 & 49.48 & 60.98 & 37.05 & 48.56 \\
    EntMin~\cite{DBLP:conf/nips/GrandvaletB04} & 46.96 & 49.59 & 61.10 & 37.26 & 48.73 \\
    FlexMatch~\cite{DBLP:conf/nips/ZhangWHWWOS21} & 38.97 & 32.44 & 39.34 & 11.82 & 30.64\\
    DebiasPL~\cite{Wang_2022_CVPR}   & 54.18 & 54.27 & 57.53 & 47.93 & 53.48\\
    FixMatch~\cite{DBLP:conf/nips/SohnBCZZRCKL20} & 51.22 & 54.99 & 59.53 & 52.45 & 54.55\\
    StyleMatch~\cite{DBLP:journals/corr/abs-2106-00592}  & 52.76 & 56.15 & 58.72 & 52.89 & 55.13\\
    MultiMatch (ours)   & \textbf{56.13} & \textbf{59.15} & \textbf{63.68} & \textbf{56.18} & \textbf{58.79}\\
    \midrule
          & \multicolumn{5}{c}{5 labels per class} \\
    \midrule
    MeanTeacher~\cite{DBLP:conf/nips/TarvainenV17} & 39.01 & 42.92 & 54.40 & 31.13 & 41.87\\
    EntMin~\cite{DBLP:conf/nips/GrandvaletB04}  & 39.39 & 43.35 & 54.80 & 31.72 & 42.32\\
    FlexMatch~\cite{DBLP:conf/nips/ZhangWHWWOS21}  & 13.80 & 20.52 & 17.88 & 20.40 & 18.15\\
     DebiasPL~\cite{Wang_2022_CVPR}  & 50.87 & 52.11 & 51.86 & 44.61 & 49.86\\
    FixMatch~\cite{DBLP:conf/nips/SohnBCZZRCKL20} & 47.23 & 52.88 & 55.99 & \textbf{52.88} & 52.24\\
    StyleMatch~\cite{DBLP:journals/corr/abs-2106-00592}  & 46.30 & 52.63 & 54.65 & 50.59 & 51.04\\
    MultiMatch (ours)  &  \textbf{50.99} & \textbf{56.03} & \textbf{58.53} & \textbf{52.88} & \textbf{54.61}\\
    \bottomrule
    \end{tabular}%
  \label{tab11}%
\end{table*}%

\subsection{Ablation Studies} \label{sec:EXP-SS}
In the experiment, we first validate the effectiveness of the multi-task learning framework, and then analyze the efficacy of the fusion prediction scheme in the training stage and the test stage, respectively. The experimental results are shown in Table~\ref{tab02}, where ``MTL+TRAIN-local+TEST-global'' denotes that the pseudo-labels are generated by the domain-specific path (\ie, local task), and the final prediction during test is based on the global task. In other words, ``MTL+TRAIN-local+TEST-global'' means that we do not utilize the fusion prediction scheme in both the training and test stages. ``MTL+TRAIN-global-local+TEST-global-local'' indicates that we employ the fusion prediction scheme in both the training and test stages.

As seen in Table~\ref{tab02}, ``MTL+TRAIN-local+TEST-global'' outperforms ``Baseline'' on both PACS and Office-Home, which confirms the effectiveness of the multi-task learning framework in the semi-supervised domain generalization task. For example, the multi-task learning framework can bring an obvious improvement by $+3.37\%$ ($61.23$ vs. $57.86$) on Office-Home. In addition, the fusion prediction scheme is also effective in both the training and test stages. As seen in Table~\ref{tab02}, ``MTL+TRAIN-global-local+TEST-global'' outperforms ``MTL+TRAIN-local+TEST-global'', and ``MTL+TRAIN-global-local+TEST-global-local'' outperforms ``MTL+TRAIN-local+TEST-global-local'', which indicates the effectiveness of the fusion prediction scheme in the training stage. Meanwhile, ``MTL+TRAIN-local+TEST-global-local '' outperforms ``MTL+TRAIN-local+TEST-global'', and ``MTL+TRAIN-global-local+TEST-global-local'' outperforms ``MTL+TRAIN-global-local+TEST-global'', which confirms the effectiveness of the fusion prediction scheme in the test stage. In our method, we use the fusion manner in Section~\ref{sec:PF}. We will also investigate some other fusion manners in further analysis.
\begin{table*}[htbp]
  \centering
  \caption{Ablation studies on different components of our method on PACS and Office-Home under the case with 10 labels per class. To better show the best result, we only use \textbf{bold} on the best average result.}
    \begin{tabular}{l|ccccc}
    \toprule
    \multicolumn{1}{c|}{\multirow{2}[4]{*}{Method}} & \multicolumn{5}{c}{PACS}        \\
\cmidrule{2-6}          & P     & A     & C     & S     & Avg   \\
    \midrule
    Baseline (FixMatch) & 87.40 & 76.85 & 69.40 & 77.68 & 77.83 \\
    MTL+TRAIN-local+TEST-global & 92.63 & 77.28 & 68.52 & 74.15 & 78.15  \\
    MTL+TRAIN-local+TEST-global-local & 93.37 & 83.43 & 70.15 & 75.71 & 80.67  \\
    MTL+TRAIN-global-local+TEST-global & 92.22 & 77.08 & 68.72 & 75.05 & 78.27  \\
    MTL+TRAIN-global-local+TEST-global-local & 93.25 & 83.30 & 73.00 & 76.74 & \textbf{81.57} \\
    \bottomrule
     \multicolumn{1}{c|}{\multirow{2}[4]{*}{Method}} & \multicolumn{5}{c}{Office-Home} \\
\cmidrule{2-6}            & A     & C     & P     & R     & Avg \\
    \midrule
    Baseline (FixMatch) & 49.90 & 50.98 & 63.79 & 66.75 & 57.86 \\
    MTL+TRAIN-local+TEST-global  & 55.04 & 51.24 & 67.41 & 71.21 & 61.23 \\
    MTL+TRAIN-local+TEST-global-local  & 56.13 & 51.70 & 68.20 & 71.71 & 61.94 \\
    MTL+TRAIN-global-local+TEST-global  & 55.31 & 52.36 & 68.17 & 71.58 & 61.86 \\
    MTL+TRAIN-global-local+TEST-global-local  & 56.32 & 52.76 & 68.94 & 72.28 & \textbf{62.57} \\
    \bottomrule
    \end{tabular}%
  \label{tab02}%
\end{table*}%

\subsection{Further Analysis}\label{sec:EXP-FA}
\textbf{Evaluation on different fusion manners.} 
In this part, we evaluate different fusion manners in our framework on PACS. Experimental results are listed in Table~\ref{tab03}. Each fusion manner is described in the following formulas. ``TEST-avg-all'' is the mean of the outputs from all tasks in the testing stage. ``TEST-max'' denotes the maximum output among all tasks in the testing stage. ``TRAIN-avg'' means using the mean of outputs from the own task and the global task during pseudo-labeling. As observed in Table~\ref{tab03}, ``TRAIN-max+TEST-avg'' (\ie, Eq.~\ref{eq06} and Eq.~\ref{eq08}) can obtain a slight improvement when compared with other fusion schemes. Besides, compared with the ``MTL+TRAIN-local+TEST-global'' (\ie, without using the fusion scheme in our method), all schemes in Table~\ref{tab03} outperform it, which indicates that the prediction's ensemble is significant for our method during training and testing.

TEST-avg-all:
     \begin{equation}
     \nonumber
  \begin{aligned}
   {\rm Class} = {\rm PredictLabel} (\frac{y_1+ \ldots +y_{N+1}}{N+1}).
  \end{aligned}
  \end{equation}

TEST-max:
     \begin{equation}
     \nonumber
  \begin{aligned}
  {\rm Class} = {\rm PredictLabel} ([y_{max}; y_{N+1}]),
  \end{aligned}
  \end{equation}
  where $y_{max}$ is defined in Eq.~\ref{eq07}.

TRAIN-avg:
 \begin{equation}
 \nonumber
  \begin{aligned}
  {\rm Class} = {\rm PredictLabel} (\frac{y_i + y_{N+1}}{2}),
  \end{aligned}
  \end{equation}
  where $y_i$ is the perdition of an image from the $i$-th domain.

\begin{table}[htbp]
  \centering
  \caption{Experimental results of different label-fusion schemes on PACS under the case with 10 labels per class.}
    \begin{tabular}{l|ccccc}
    \toprule
    \multicolumn{1}{c|}{Fusion scheme} & P     & A     & C     & S     & Avg \\
    \midrule
    TRAIN-max+TEST-avg  (ours) & 93.25 & 83.30 & 73.00 & 76.74 & \textbf{81.57} \\
    TRAIN-max+TEST-avg-all & 93.36 & 83.32 & 72.34 & 76.76 & 81.44 \\
    TRAIN-max+TEST-max & 92.93 & 83.24 & 72.76 & 76.76 & 81.42 \\
    \midrule
    TRAIN-avg+TEST-avg & 93.45 & 83.56 & 71.26 & 76.89 & 81.29 \\
    TRAIN-avg+TEST-avg-all & 93.46 & 83.49 & 71.35 & 76.82 & 81.28 \\
    TRAIN-avg+TEST-max & 93.29 & 83.59 & 71.08 & 76.96 & 81.23 \\
    \bottomrule
    \end{tabular}%
  \label{tab03}%
\end{table}%

\begin{figure}
\centering
\subfigure[PACS]{
\includegraphics[width=5.51cm]{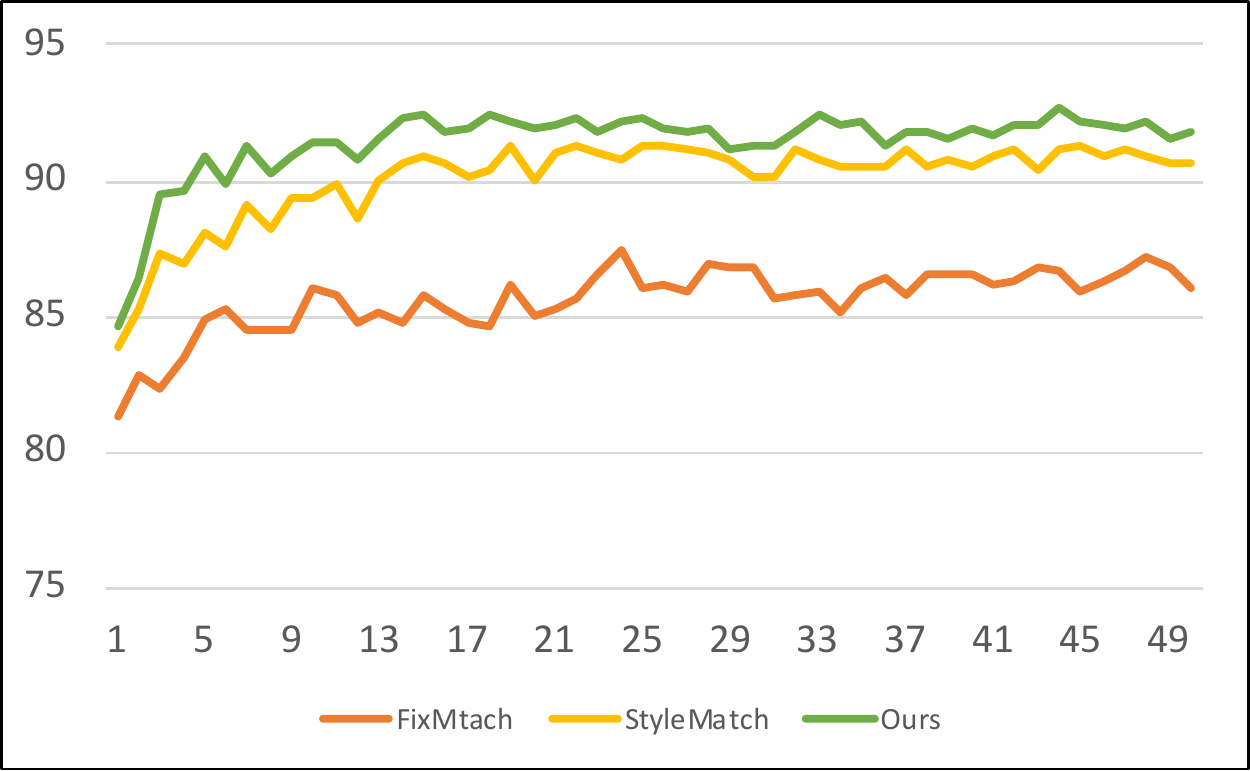}
}
\subfigure[Office-Home]{
\includegraphics[width=5.51cm]{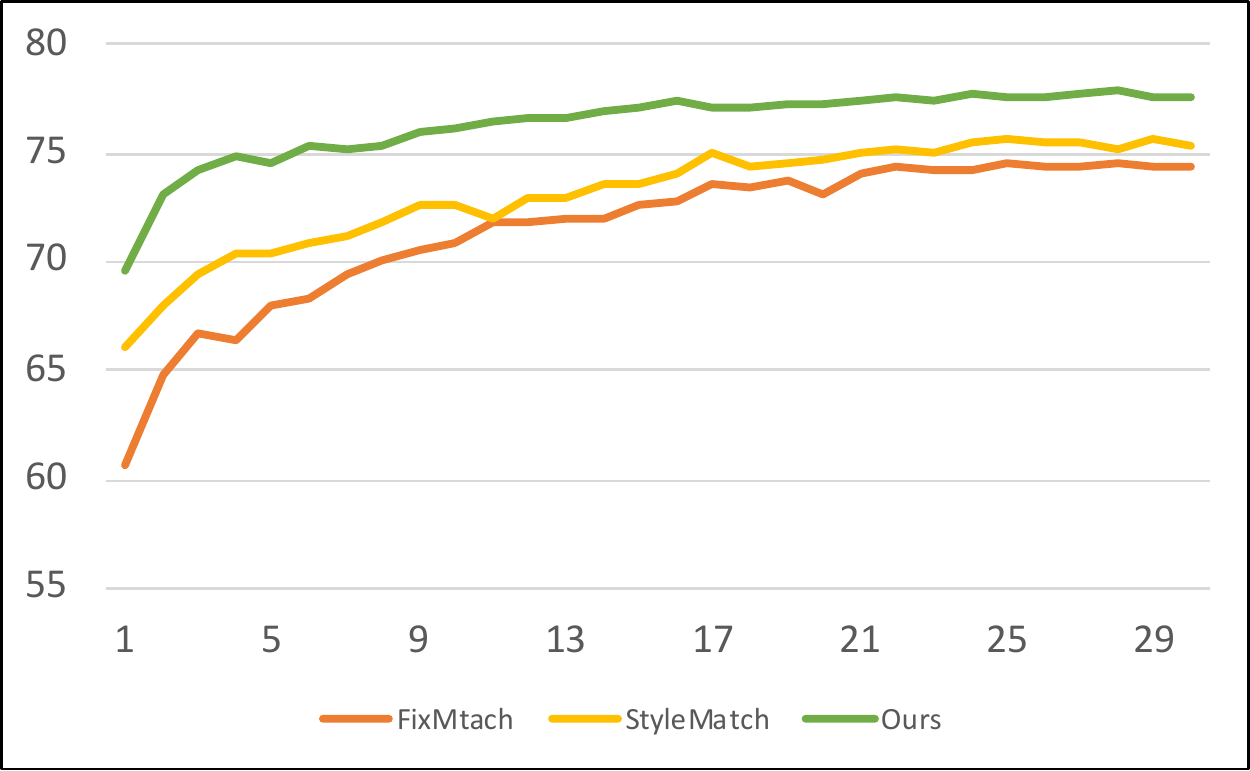}
}
\caption{The precision of pseudo-labels of different methods at different epochs on PACS and Office-Home, respectively.}
\label{fig02}
\end{figure}

\textbf{The accuracy of pseudo-labels.} In this experiment,  we use \textbf{Precision}, \textbf{Recall} and \textbf{macro-f1} to evaluate the accuracy of pseudo-labels. Here we give the detailed definition as follows:
  \begin{equation}
  \begin{aligned}
  \textbf{Precision} = \frac{N_c}{N}\times 100,
  \end{aligned}
    \label{eq23}
  \end{equation}
  where $N$ and $N_c$ denote the number of the total samples and the correctly predicted samples.
Besides, we define $TP$: true positive. $TN$: true negative. $FP$: false positive. $FN$: false negative. To be specific, for the $i$-th class,\\ 
$TP_i$ is the number of the true $i$-th class among the predicted $i$-th class;\\
$TN_i$ is the number of the true non $i$-th class among the predicted non $i$-th class; \\
$FP_i$ is the number of the false $i$-th class among the predicted $i$-th class; \\
$FN_i$ is the number of the false non $i$-th class among the predicted non $i$-th class.\\
For the $i$-th class,  we have
    \begin{equation}
  \begin{aligned}
  Recall_i = \frac{TP_i}{TP_i+FN_i},~~~ Precision_i= \frac{TP_i}{TP_i+FP_i}, ~~~ F_i=2\frac{Precision_i\times Recall_i}{Precision_i+ Recall_i}.
  \end{aligned}
    \label{eqR21}
  \end{equation}
Hence, we can obtain:
    \begin{equation}
  \begin{aligned}
   \textbf{Recall} = \frac{100}{C}\sum_{c=1}^{C}{Recall_i}, ~~~~~~~~~~~ \textbf{macro-f1} = \frac{100}{C}\sum_{c=1}^{C}{F_i},
  \end{aligned}
    \label{eqR22}
  \end{equation}
  where $C$ is the number of classes.
Table~\ref{tab04} shows the experimental results, where ``MTL+TRAIN-local'' represents the accuracy of the pseudo-labels from the domain-specific classifier in our method, and ``MTL+TRAIN-global-local'' is the accuracy of the pseudo-labels generated by fusing the domain-specific classifier and the global classifier in our method. As observed in Table~\ref{tab04}, ``MTL+TRAIN-local'' can improve the macro-f1 of  FixMatch by  $+1.37$ ($90.63$ vs. $89.26$) and $+1.27$ ($70.41$ vs. $69.14$) on PACS and Office-Home, respectively. This validates that using the independent task for each domain can indeed alleviate the interference of different domains so as to improve the pseudo-labels of unlabeled data. Besides, ``MTL+TRAIN-global-local'' outperforms all other methods, \eg, ``MTL+TRAIN-global-local'' increases the macro-f1 of StyleMatch by $+1.35$ ($92.70$ vs. $91.35$) and $+2.39$ ($73.07$ vs. $70.68$) on on PACS and Office-Home, respectively. Therefore, this confirms our method can achieve better accuracy of pseudo-labels to enhance the generalization capability of the model. Furthermore, we also display the accuracy of pseudo-labels of different methods at different epochs on PACS and Office-Home in Fig.~\ref{fig02}, respectively. As seen, our method can give more accurate pseudo-labels at each epoch when compared with FixMatch and StyleMatch.
\begin{table}[htbp]
  \centering
  \caption{The accuracy of pseudo-labels of different methods on PACS and Office-Home under the case with 10 labels per class.}
    \begin{tabular}{c|l|ccc}
    \toprule
    Dataset & \multicolumn{1}{c|}{Method} & Precision & Recall & macro-f1 \\
    \midrule
    \multirow{4}[2]{*}{PACS} & FixMatch & 88.83 & 89.52 & 89.26 \\
          & StyleMatch & 90.84 & 91.52 & 91.35 \\
          & MTL+TRAIN-local & 90.17 & 90.86 & 90.63 \\
          & MTL+TRAIN-global-local & \textbf{92.23} & \textbf{92.82} & \textbf{92.70} \\
    \midrule
    \multirow{4}[2]{*}{Office-Home} & FixMatch & 71.05 & 69.97 & 69.14 \\
          & StyleMatch & 72.67 & 71.41 & 70.68 \\
          & MTL+TRAIN-local & 72.22 & 71.25 & 70.41 \\
          & MTL+TRAIN-global-local & \textbf{74.86} & \textbf{73.76} & \textbf{73.07} \\
    \bottomrule
    \end{tabular}%
  \label{tab04}%
\end{table}%

\begin{figure}[t]
\centering
\includegraphics[width=10cm]{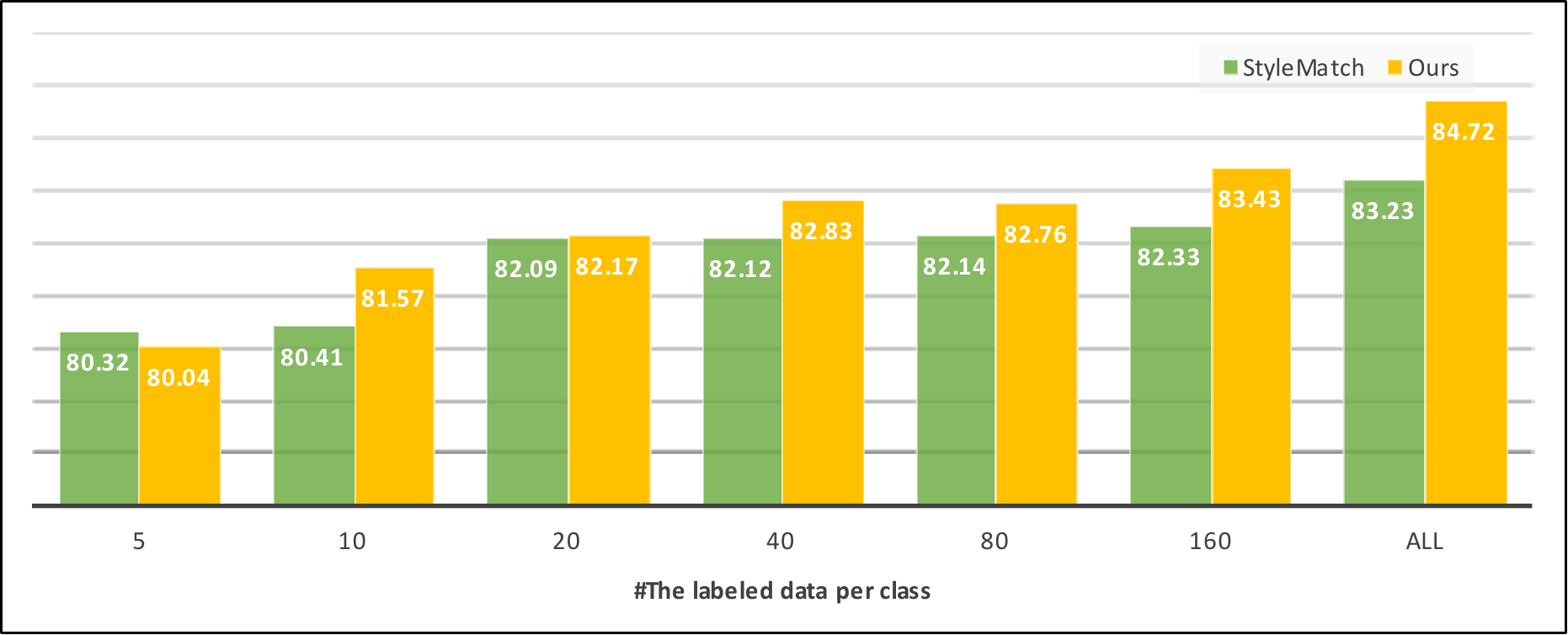}
\caption{Comparison between our MultiMatch and StyleMatch under different numbers of the labeled data on PACS. ``ALL'' denotes that all training samples have label information during training.}
\label{fig03}
\end{figure}

\textbf{Test on different numbers of the labeled data.}
We also conduct the comparison between our proposed MultiMatch and StyleMatch under different numbers of the labeled data on PACS, as shown in Fig.~\ref{fig03}. According to this figure, except for the ``5 labels per class'' case, our MultiMatch obviously outperforms StyleMatch in all other cases. For example, when we use 160 labels per class, our MultiMatch improves the performance by $+1.1\%$ ($83.43$ vs. $82.33$) when compared with StyleMatch. Besides, given the label information for all training samples, our method can also obtain better results than StyleMatch.

\textbf{Evaluation on the independent BN and the independent classifier.} In this part, we validate the effectiveness of the independent BN and the independent classifier in our method. Experimental results are listed in Table~\ref{tab07}, where ``w/ SBN'' and ``w/ SC'' indicate using the shared BN and the shared classifier in our method, respectively. As observed in Table~\ref{tab07}, using the independent BN and the independent classifier together can yield better performance than using the independent BN or the independent classifier. For example, our method improve the ``w/ SBN'' and ``w/ SC'' by $+1.44\%$ ($81.57$ vs. $80.13$) and $+2.47\%$ ($81.57$ vs. $79.10$) on PACS, respectively. Besides, compared with the original FixMatch in Table~\ref{tab01} under ``10 labels per class'', both ``w/ SBN'' and ``w/ SC'' outperform them on PACS and Office-Home. All the above observations confirm the effectiveness of both the independent BN and the independent classifier in our method.
\begin{table}[htbp]
  \centering
  \caption{Evaluation on the independent BN and the independent classifier on PACS and Office-Home, respectively.}
    \begin{tabular}{l|ccccc}
    \toprule
    \multicolumn{1}{c|}{\multirow{2}[4]{*}{Module}} & \multicolumn{5}{c}{PACS} \\
\cmidrule{2-6}          & P     & A     & C     & S     & Avg \\
    \midrule
    w/ SBN & ~~91.62~~ & ~~82.09~~ & ~~73.10~~ & ~~73.70~~ & ~~80.13~~ \\
    w/ SC & 91.07 & 78.10 & 71.12 & 76.12 & 79.10 \\
    Ours  & 93.25 & 83.30 & 73.00 & 76.74 & \textbf{81.57} \\
    \midrule
    \multicolumn{1}{c|}{\multirow{2}[4]{*}{Module}} & \multicolumn{5}{c}{Office-Home} \\
\cmidrule{2-6}          & A     & C     & P     & R     & Avg \\
    \midrule
    w/ SBN & 55.30 & 51.65 & 68.01 & 71.56 & 61.63 \\
    w/ SC & 54.68 & 52.28 & 67.38 & 71.22 & 61.39 \\
    Ours  & 56.32 & 52.76 & 68.94 & 72.28 & \textbf{62.57} \\
    \bottomrule
    \end{tabular}%
  \label{tab07}%
\end{table}%

\begin{figure}
\centering
\subfigure[PACS]{
\includegraphics[width=5.51cm]{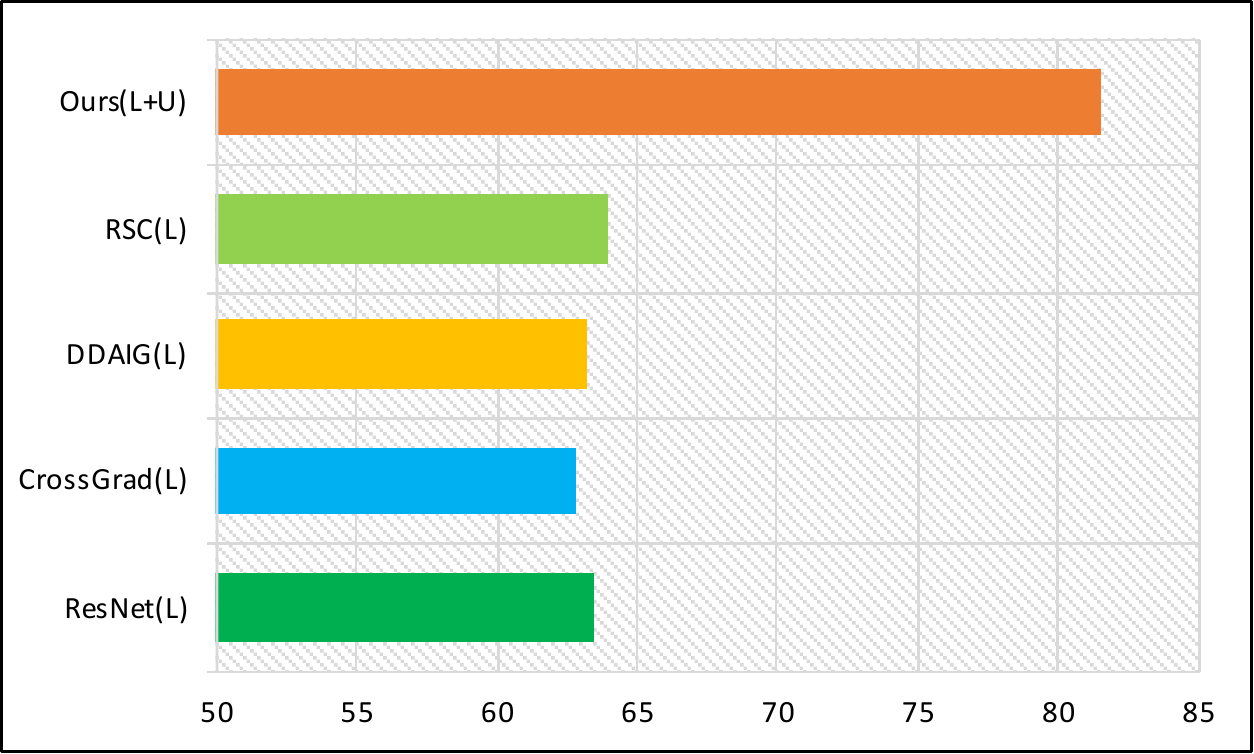}
}
\subfigure[Office-Home]{
\includegraphics[width=5.51cm]{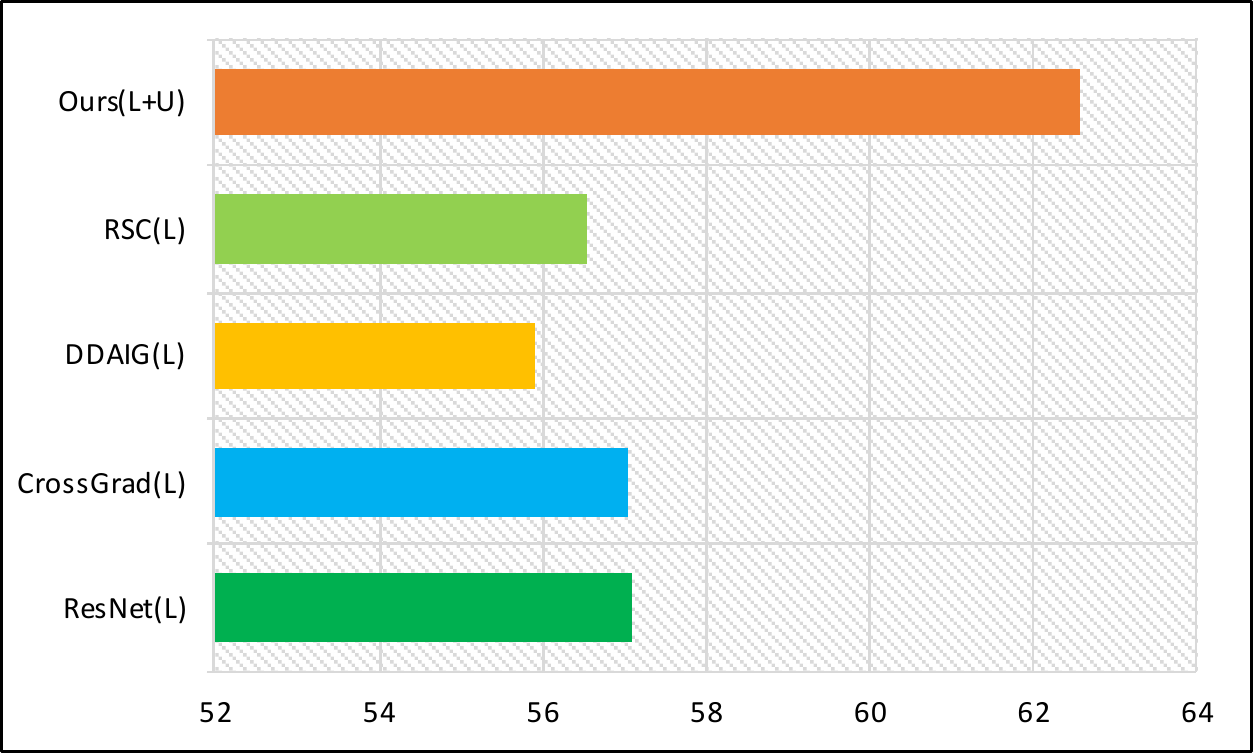}
}
\caption{Experimental results of the typical DG methods under the labeled samples in SSDG on PACS and Office-Home, respectively. Note that ``L'' denotes only using the labeled samples, and ``L+U'' uses the labeled and unlabeled samples together.}
\label{fig06}
\end{figure}

\textbf{Effectiveness of the unlabeled data in SSDG.}
To validate the effectiveness of the unlabeled data in SSDG, we train the supervised DG methods using the labeled samples, including ResNet18, CrossGrad~\cite{DBLP:conf/iclr/ShankarPCCJS18}, DDAIG~\cite{DBLP:conf/aaai/ZhouYHX20}, RSC~\cite{DBLP:conf/eccv/HuangWXH20}. Experimental results are shown in Fig.~\ref{fig06}. As observed in this figure, using the unlabeled data can obtain better performance than these conventional DG methods with the labeled data, which validates that the unlabeled samples are very meaningful in the SSDG task. Furthermore, our MultiMatch can effectively mine the information from these unlabeled data, which has been reported in Table~\ref{tab01}. 

\textbf{Test on the supervised case.} In the experiment, we train our model in the supervised setting, and compare it with some supervised DG methods, as reported in Tables~\ref{tab05} and \ref{tab06}. MLDG~\cite{DBLP:conf/aaai/LiYSH18}, MASF~\cite{DBLP:conf/nips/DouCKG19} and MetaReg~\cite{DBLP:conf/nips/BalajiSC18} are meta-learning based methods. FACT~\cite{DBLP:conf/cvpr/XuZ0W021}, RSC~\cite{DBLP:conf/eccv/HuangWXH20} and FSDCL~\cite{DBLP:conf/mm/JeonHLLB21} are augmentation based methods. 
In addition, VDN~\cite{wang2021variational} and SNR~\cite{DBLP:journals/tmm/JinLZC22} aim at learning domain-invariant features, and  DAEL~\cite{DBLP:journals/tip/ZhouYQX21} is an ensemble learning based method. Compared with these supervised methods, our MultiMatch is also competitive in the supervised case.
\begin{table}[htbp]
  \centering
  \caption{Experimental results of different methods under the supervised setting on PACS. The  \textbf{bold} and \textbf{\color{gray}{gray}} are the best result and the second-best result, respectively.}
    \begin{tabular}{l|ccccc}
    \toprule
    \multicolumn{1}{c|}{Method} & P     & A     & C     & S     & Avg \\
    \midrule
    MLDG~\cite{DBLP:conf/aaai/LiYSH18}  & 94.30 & 79.50 & 77.30 & 71.50 & 80.65 \\
    MASF~\cite{DBLP:conf/nips/DouCKG19}  & 94.99 & 80.29 & 77.17 & 71.69 & 81.04 \\
    MetaReg~\cite{DBLP:conf/nips/BalajiSC18} & 95.50 & 83.70 & 77.20 & 70.30 & 81.68 \\
    SNR~\cite{DBLP:journals/tmm/JinLZC22} & 94.50 & 80.3 & 78.20 & 74.10 & 81.80 \\
    VDN~\cite{wang2021variational}   & 94.00 & 82.60 & 78.50 & 82.70 & 84.45 \\
    FACT~\cite{DBLP:conf/cvpr/XuZ0W021}  & 95.15 & 85.37 & 78.38 & 79.15 & \textbf{\color{gray}{84.51}} \\
    MultiMatch (ours) & 95.79 & 84.44 & 75.83 & 82.82 & \textbf{84.72} \\
    \bottomrule
    \end{tabular}%
  \label{tab05}%
\end{table}%

\begin{table}[htbp]
  \centering
  \caption{Experimental results of different methods under the supervised setting on Office-Home.}
    \begin{tabular}{l|ccccc}
    \toprule
    \multicolumn{1}{c|}{Method} & A     & C     & P     & R     & Avg \\
    \midrule
    RSC~\cite{DBLP:conf/eccv/HuangWXH20}   & 58.42 & 47.90 & 71.63 & 74.54 & 63.12 \\
    DAEL~\cite{DBLP:journals/tip/ZhouYQX21}  & 59.40 & 55.10 & 74.00 & 75.70 & 66.05 \\
    SNR~\cite{DBLP:journals/tmm/JinLZC22} & 61.20 & 53.70 & 74.20 & 75.10 & 66.10 \\
    FSDCL~\cite{DBLP:conf/mm/JeonHLLB21} & 60.24 & 53.54 & 74.36 & 76.66 & 66.20 \\
    FACT~\cite{DBLP:conf/cvpr/XuZ0W021}  & 60.34 & 54.85 & 74.48 & 76.55 & \textbf{66.56} \\
    MultiMatch (ours)  & 59.71 & 56.25 & 74.28 & 75.70 & \textbf{\color{gray}{66.49}} \\
    \bottomrule
    \end{tabular}%
  \label{tab06}%
\end{table}%

\textbf{Comparison with ``local-to-global'' method.} In our task, we achieve the scheme in~\cite{isobe2021multi} that uses each local branch prediction to distill global branch prediction. Experimental results are shown in Tab.~\ref{tabR02}. As seen in this table, our method can obtain the better performance than the L2G~\cite{isobe2021multi}. The advantage of our method is fusing the prediction of global task based on our ``PredictLabel''  scheme in Function~\ref{f01} can further improve the accuracy of the pseudo-label, which has been validated in Tab.~\ref{tab04} of our paper.

\begin{table}[htbp]
  \centering
  \caption{comparison between our method and L2G~\cite{isobe2021multi} on PACS and Office-Home under 10 labels per class.}
    \begin{tabular}{l|ccccc}
    \toprule
    \multicolumn{6}{c}{PACS} \\
    \midrule
    \multicolumn{1}{c|}{Method} & P     & A     & C     & S     & Avg \\
    \midrule
    L2G~\cite{isobe2021multi}   & \textbf{93.40} & 83.24 & 69.18 & 75.70 & 80.38 \\
    Ours  & 93.25 & \textbf{83.30} & \textbf{73.00} & \textbf{76.74} & \textbf{81.57} \\
    \midrule
    \multicolumn{6}{c}{Office-Home} \\
    \midrule
    \multicolumn{1}{c|}{Method} & A     & C     & P     & R     & Avg \\
    \midrule
    L2G~\cite{isobe2021multi}   & 55.92 & 51.58 & 68.12 & 71.62 & 61.81 \\
    Ours  & \textbf{56.32} & \textbf{52.76} & \textbf{68.94} & \textbf{72.28} & \textbf{62.57} \\
    \bottomrule
    \end{tabular}%
  \label{tabR02}%
\end{table}%

\section{Conclusion}\label{s-conclusion}
In this paper, we aim to tackle the semi-supervised domain generalization (SSDG) task. Different from the typical semi-supervised task, the challenge of SSDG is that there exist multiple different domains with latent distribution discrepancy. To address this issue, we first explore the theory of multi-domain learning to generate more accurate pseudo-labels for unlabeled samples. Then, we propose to utilize a multi-task learning framework to mitigate the impact of the domain discrepancy and sufficiently exploit all training samples, which can effectively enhance the model's generalization. We conduct the experiment on multiple benchmark datasets, which verifies the effectiveness of the proposed method.

{\small
\bibliographystyle{acm}
\bibliography{sigproc}

\begin{thebibliography}{10}

\bibitem{DBLP:conf/nips/BalajiSC18}
{\sc Balaji, Y., Sankaranarayanan, S., and Chellappa, R.}
\newblock Metareg: Towards domain generalization using meta-regularization.
\newblock In {\em Advances in Neural Information Processing Systems
  (NeurIPS)\/} (2018), pp.~1006--1016.

\bibitem{ben2010theory}
{\sc Ben-David, S., Blitzer, J., Crammer, K., Kulesza, A., Pereira, F., and
  Vaughan, J.~W.}
\newblock A theory of learning from different domains.
\newblock {\em Machine Learning (ML) 79}, 1 (2010), 151--175.

\bibitem{DBLP:conf/iclr/BerthelotCCKSZR20}
{\sc Berthelot, D., Carlini, N., Cubuk, E.~D., Kurakin, A., Sohn, K., Zhang,
  H., and Raffel, C.}
\newblock Remixmatch: Semi-supervised learning with distribution matching and
  augmentation anchoring.
\newblock In {\em International Conference on Learning Representations
  (ICLR)\/} (2020).

\bibitem{DBLP:conf/nips/BerthelotCGPOR19}
{\sc Berthelot, D., Carlini, N., Goodfellow, I.~J., Papernot, N., Oliver, A.,
  and Raffel, C.}
\newblock Mixmatch: {A} holistic approach to semi-supervised learning.
\newblock In {\em Advances in Neural Information Processing Systems
  (NeurIPS)\/} (2019), pp.~5050--5060.

\bibitem{DBLP:conf/cvpr/CarlucciDBCT19}
{\sc Carlucci, F.~M., D'Innocente, A., Bucci, S., Caputo, B., and Tommasi, T.}
\newblock Domain generalization by solving jigsaw puzzles.
\newblock In {\em IEEE Conference on Computer Vision and Pattern Recognition
  (CVPR)\/} (2019), pp.~2229--2238.

\bibitem{DBLP:conf/cvpr/ChangYSKH19}
{\sc Chang, W., You, T., Seo, S., Kwak, S., and Han, B.}
\newblock Domain-specific batch normalization for unsupervised domain
  adaptation.
\newblock In {\em {IEEE} Conference on Computer Vision and Pattern Recognition
  (CVPR)\/} (2019), pp.~7354--7362.

\bibitem{DBLP:conf/cvpr/DengDSLL009}
{\sc Deng, J., Dong, W., Socher, R., Li, L., Li, K., and Li, F.}
\newblock Imagenet: {A} large-scale hierarchical image database.
\newblock In {\em IEEE Conference on Computer Vision and Pattern Recognition
  (CVPR)\/} (2009), pp.~248--255.

\bibitem{devlin2018bert}
{\sc Devlin, J., Chang, M.-W., Lee, K., and Toutanova, K.}
\newblock Bert: Pre-training of deep bidirectional transformers for language
  understanding.
\newblock {\em arXiv preprint arXiv:1810.04805\/} (2018).

\bibitem{DBLP:conf/nips/DouCKG19}
{\sc Dou, Q., de~Castro, D.~C., Kamnitsas, K., and Glocker, B.}
\newblock Domain generalization via model-agnostic learning of semantic
  features.
\newblock In {\em Advances in Neural Information Processing Systems
  (NeurIPS)\/} (2019), pp.~6447--6458.

\bibitem{fu2021dynamic}
{\sc Fu, S., Liu, W., Guan, W., Zhou, Y., Tao, D., and Xu, C.}
\newblock Dynamic graph learning convolutional networks for semi-supervised
  classification.
\newblock {\em ACM Transactions on Multimedia Computing, Communications, and
  Applications (TOMM) 17}, 1s (2021), 1--13.

\bibitem{Gong_2021_CVPR}
{\sc Gong, C., Wang, D., and Liu, Q.}
\newblock Alphamatch: Improving consistency for semi-supervised learning with
  alpha-divergence.
\newblock In {\em IEEE Conference on Computer Vision and Pattern Recognition
  (CVPR)\/} (2021), pp.~13683--13692.

\bibitem{DBLP:conf/cvpr/GongLCG19}
{\sc Gong, R., Li, W., Chen, Y., and Gool, L.~V.}
\newblock {DLOW:} domain flow for adaptation and generalization.
\newblock In {\em IEEE Conference on Computer Vision and Pattern Recognition
  (CVPR)\/} (2019), pp.~2477--2486.

\bibitem{DBLP:conf/nips/GrandvaletB04}
{\sc Grandvalet, Y., and Bengio, Y.}
\newblock Semi-supervised learning by entropy minimization.
\newblock In {\em Advances in Neural Information Processing Systems
  (NeurIPS)\/} (2004), pp.~529--536.

\bibitem{hastie2009elements}
{\sc Hastie, T., Tibshirani, R., Friedman, J.~H., and Friedman, J.~H.}
\newblock {\em The elements of statistical learning: data mining, inference,
  and prediction}, vol.~2.
\newblock 2009.

\bibitem{DBLP:conf/cvpr/HeZRS16}
{\sc He, K., Zhang, X., Ren, S., and Sun, J.}
\newblock Deep residual learning for image recognition.
\newblock In {\em IEEE Conference on Computer Vision and Pattern Recognition
  (CVPR)\/} (2016), pp.~770--778.

\bibitem{DBLP:conf/iccv/HuangB17}
{\sc Huang, X., and Belongie, S.~J.}
\newblock Arbitrary style transfer in real-time with adaptive instance
  normalization.
\newblock In {\em International Conference on Computer Vision (ICCV)\/} (2017),
  pp.~1510--1519.

\bibitem{DBLP:conf/eccv/HuangWXH20}
{\sc Huang, Z., Wang, H., Xing, E.~P., and Huang, D.}
\newblock Self-challenging improves cross-domain generalization.
\newblock In {\em European Conference on Computer Vision (ECCV)\/} (2020),
  pp.~124--140.

\bibitem{isobe2021multi}
{\sc Isobe, T., Jia, X., Chen, S., He, J., Shi, Y., Liu, J., Lu, H., and Wang,
  S.}
\newblock Multi-target domain adaptation with collaborative consistency
  learning.
\newblock In {\em Proceedings of the IEEE/CVF Conference on Computer Vision and
  Pattern Recognition (CVPR)\/} (2021), pp.~8187--8196.

\bibitem{DBLP:conf/mm/JeonHLLB21}
{\sc Jeon, S., Hong, K., Lee, P., Lee, J., and Byun, H.}
\newblock Feature stylization and domain-aware contrastive learning for domain
  generalization.
\newblock In {\em ACM International Conference on Multimedia (MM)\/} (2021),
  pp.~22--31.

\bibitem{DBLP:journals/tmm/JinLZC22}
{\sc Jin, X., Lan, C., Zeng, W., and Chen, Z.}
\newblock Style normalization and restitution for domain generalization and
  adaptation.
\newblock {\em IEEE Transactions on Multimedia (TMM) 24\/} (2022), 3636--3651.

\bibitem{DBLP:conf/iclr/LaineA17}
{\sc Laine, S., and Aila, T.}
\newblock Temporal ensembling for semi-supervised learning.
\newblock In {\em International Conference on Learning Representations
  (ICLR)\/} (2017).

\bibitem{DBLP:conf/aaai/LiYSH18}
{\sc Li, D., Yang, Y., Song, Y., and Hospedales, T.~M.}
\newblock Learning to generalize: Meta-learning for domain generalization.
\newblock In {\em AAAI Conference on Artificial Intelligence (AAAI)\/} (2018),
  pp.~3490--3497.

\bibitem{Li2017DeeperBA}
{\sc Li, D., Yang, Y., Song, Y.-Z., and Hospedales, T.~M.}
\newblock Deeper, broader and artier domain generalization.
\newblock In {\em International Conference on Computer Vision (ICCV)\/} (2017),
  pp.~5543--5551.

\bibitem{DBLP:conf/iccv/LiZYLSH19}
{\sc Li, D., Zhang, J., Yang, Y., Liu, C., Song, Y., and Hospedales, T.~M.}
\newblock Episodic training for domain generalization.
\newblock In {\em International Conference on Computer Vision (ICCV)\/} (2019),
  pp.~1446--1455.

\bibitem{DBLP:conf/cvpr/LiPWK18}
{\sc Li, H., Pan, S.~J., Wang, S., and Kot, A.~C.}
\newblock Domain generalization with adversarial feature learning.
\newblock In {\em IEEE Conference on Computer Vision and Pattern Recognition
  (CVPR)\/} (2018), pp.~5400--5409.

\bibitem{Li_2021_ICCV_Comatch}
{\sc Li, J., Xiong, C., and Hoi, S.~C.}
\newblock Comatch: Semi-supervised learning with contrastive graph
  regularization.
\newblock In {\em International Conference on Computer Vision (ICCV)\/} (2021),
  pp.~9475--9484.

\bibitem{DBLP:conf/eccv/LiTGLLZT18}
{\sc Li, Y., Tian, X., Gong, M., Liu, Y., Liu, T., Zhang, K., and Tao, D.}
\newblock Deep domain generalization via conditional invariant adversarial
  networks.
\newblock In {\em European Conference on Computer Vision (ECCV)\/} (2018),
  pp.~647--663.

\bibitem{liu2022category}
{\sc Liu, Y., Xiong, Z., Li, Y., Lu, Y., Tian, X., and Zha, Z.-J.}
\newblock Category-stitch learning for union domain generalization.
\newblock {\em ACM Transactions on Multimedia Computing, Communications, and
  Applications (TOMM)\/} (2022).

\bibitem{liu2021dg}
{\sc Liu, Y., Xiong, Z., Li, Y., Tian, X., and Zha, Z.-J.}
\newblock Domain generalization via encoding and resampling in a unified latent
  space.
\newblock {\em IEEE Transactions on Multimedia (TMM)\/} (2021).

\bibitem{DBLP:journals/tmm/LuoJGLLLG20}
{\sc Luo, H., Jiang, W., Gu, Y., Liu, F., Liao, X., Lai, S., and Gu, J.}
\newblock A strong baseline and batch normalization neck for deep person
  re-identification.
\newblock {\em IEEE Transactions on Multimedia (TMM) 22}, 10 (2020),
  2597--2609.

\bibitem{DBLP:conf/icml/MuandetBS13}
{\sc Muandet, K., Balduzzi, D., and Sch{\"{o}}lkopf, B.}
\newblock Domain generalization via invariant feature representation.
\newblock In {\em International Conference on Machine Learning (ICML)\/}
  (2013), pp.~10--18.

\bibitem{DBLP:conf/cvpr/NamLPYY21}
{\sc Nam, H., Lee, H., Park, J., Yoon, W., and Yoo, D.}
\newblock Reducing domain gap by reducing style bias.
\newblock In {\em IEEE Conference on Computer Vision and Pattern Recognition
  (CVPR)\/} (2021), pp.~8690--8699.

\bibitem{Nassar_2021_CVPR}
{\sc Nassar, I., Herath, S., Abbasnejad, E., Buntine, W., and Haffari, G.}
\newblock All labels are not created equal: Enhancing semi-supervision via
  label grouping and co-training.
\newblock In {\em IEEE Conference on Computer Vision and Pattern Recognition
  (CVPR)\/} (2021), pp.~7241--7250.

\bibitem{Oh_2022_CVPR}
{\sc Oh, Y., Kim, D.-J., and Kweon, I.~S.}
\newblock Daso: Distribution-aware semantics-oriented pseudo-label for
  imbalanced semi-supervised learning.
\newblock In {\em IEEE Conference on Computer Vision and Pattern Recognition
  (CVPR)\/} (2022), pp.~9786--9796.

\bibitem{DBLP:conf/iccv/PengBXHSW19}
{\sc Peng, X., Bai, Q., Xia, X., Huang, Z., Saenko, K., and Wang, B.}
\newblock Moment matching for multi-source domain adaptation.
\newblock In {\em International Conference on Computer Vision (ICCV)\/} (2019),
  pp.~1406--1415.

\bibitem{DBLP:journals/tomccap/QiWHSG21}
{\sc Qi, L., Wang, L., Huo, J., Shi, Y., and Gao, Y.}
\newblock Greyreid: {A} novel two-stream deep framework with rgb-grey
  information for person re-identification.
\newblock {\em ACM Transactions on Multimedia Computing, Communications, and
  Applications (TOMM) 17}, 1 (2021), 27:1--27:22.

\bibitem{qi2022novel}
{\sc Qi, L., Wang, L., Shi, Y., and Geng, X.}
\newblock A novel mix-normalization method for generalizable multi-source
  person re-identification.
\newblock {\em IEEE Transactions on Multimedia (TMM)\/} (2022).

\bibitem{DBLP:conf/wacv/RahmanFBS19}
{\sc Rahman, M.~M., Fookes, C., Baktashmotlagh, M., and Sridharan, S.}
\newblock Multi-component image translation for deep domain generalization.
\newblock In {\em IEEE Winter Conference on Applications of Computer Vision
  (WACV)\/} (2019), pp.~579--588.

\bibitem{DBLP:journals/pr/RahmanFBS20}
{\sc Rahman, M.~M., Fookes, C., Baktashmotlagh, M., and Sridharan, S.}
\newblock Correlation-aware adversarial domain adaptation and generalization.
\newblock {\em Pattern Recognition (PR) 100\/} (2020), 107124.

\bibitem{DBLP:journals/pami/RenHG017}
{\sc Ren, S., He, K., Girshick, R.~B., and Sun, J.}
\newblock Faster {R-CNN:} towards real-time object detection with region
  proposal networks.
\newblock {\em IEEE Transactions on Pattern Analysis and Machine Intelligence
  (TPAMI) 39}, 6 (2017), 1137--1149.

\bibitem{DBLP:conf/eccv/SeoSKKHH20}
{\sc Seo, S., Suh, Y., Kim, D., Kim, G., Han, J., and Han, B.}
\newblock Learning to optimize domain specific normalization for domain
  generalization.
\newblock In {\em European Conference on Computer Vision (ECCV)\/} (2020),
  pp.~68--83.

\bibitem{DBLP:conf/iclr/ShankarPCCJS18}
{\sc Shankar, S., Piratla, V., Chakrabarti, S., Chaudhuri, S., Jyothi, P., and
  Sarawagi, S.}
\newblock Generalizing across domains via cross-gradient training.
\newblock In {\em International Conference on Learning Representations
  (ICLR)\/} (2018).

\bibitem{DBLP:conf/nips/SohnBCZZRCKL20}
{\sc Sohn, K., Berthelot, D., Carlini, N., Zhang, Z., Zhang, H., Raffel, C.,
  Cubuk, E.~D., Kurakin, A., and Li, C.}
\newblock Fixmatch: Simplifying semi-supervised learning with consistency and
  confidence.
\newblock In {\em Advances in Neural Information Processing Systems
  (NeurIPS)\/} (2020).

\bibitem{DBLP:conf/nips/TarvainenV17}
{\sc Tarvainen, A., and Valpola, H.}
\newblock Mean teachers are better role models: Weight-averaged consistency
  targets improve semi-supervised deep learning results.
\newblock In {\em Advances in Neural Information Processing Systems
  (NeurIPS)\/} (2017), pp.~1195--1204.

\bibitem{van2008visualizing}
{\sc Van~der Maaten, L., and Hinton, G.}
\newblock Visualizing data using t-sne.
\newblock {\em Journal of machine learning research (JMLR) 9}, 11 (2008),
  2579--2605.

\bibitem{venkateswara2017deep}
{\sc Venkateswara, H., Eusebio, J., Chakraborty, S., and Panchanathan, S.}
\newblock Deep hashing network for unsupervised domain adaptation.
\newblock In {\em CVPR\/} (2017), pp.~5018--5027.

\bibitem{DBLP:conf/ijcai/0001LLOQ21}
{\sc Wang, J., Lan, C., Liu, C., Ouyang, Y., and Qin, T.}
\newblock Generalizing to unseen domains: {A} survey on domain generalization.
\newblock In {\em International Joint Conference on Artificial Intelligence
  (IJCAI)\/} (2021), pp.~4627--4635.

\bibitem{Wang_2022_CVPR}
{\sc Wang, X., Wu, Z., Lian, L., and Yu, S.~X.}
\newblock Debiased learning from naturally imbalanced pseudo-labels.
\newblock In {\em IEEE Conference on Computer Vision and Pattern Recognition
  (CVPR)\/} (2022), pp.~14647--14657.

\bibitem{wang2021pointwise}
{\sc Wang, Y., Han, J., Shen, Y., and Xue, H.}
\newblock Pointwise manifold regularization for semi-supervised learning.
\newblock {\em Frontiers of Computer Science (FCS) 15\/} (2021), 1--8.

\bibitem{wang2021variational}
{\sc Wang, Y., Li, H., Chau, L.-P., and Kot, A.~C.}
\newblock Variational disentanglement for domain generalization.
\newblock {\em arXiv preprint arXiv:2109.05826\/} (2021).

\bibitem{wang2022feature}
{\sc Wang, Y., Qi, L., Shi, Y., and Gao, Y.}
\newblock Feature-based style randomization for domain generalization.
\newblock {\em IEEE Transactions on Circuits and Systems for Video Technology
  (TCSVT) 32}, 8 (2022), 5495--5509.

\bibitem{wu2023domain}
{\sc Wu, K., Jia, F., and Han, Y.}
\newblock Domain-specific feature elimination: multi-source domain adaptation
  for image classification.
\newblock {\em Frontiers of Computer Science (FCS) 17}, 4 (2023), 174705.

\bibitem{DBLP:conf/cvpr/XuZ0W021}
{\sc Xu, Q., Zhang, R., Zhang, Y., Wang, Y., and Tian, Q.}
\newblock A fourier-based framework for domain generalization.
\newblock In {\em IEEE Conference on Computer Vision and Pattern Recognition
  (CVPR)\/} (2021), pp.~14383--14392.

\bibitem{DBLP:journals/tomccap/XuSDWXH22}
{\sc Xu, Y., Sheng, K., Dong, W., Wu, B., Xu, C., and Hu, B.}
\newblock Towards corruption-agnostic robust domain adaptation.
\newblock {\em ACM Transactions on Multimedia Computing, Communications, and
  Applications (TOMM) 18}, 4 (2022), 99:1--99:16.

\bibitem{Yang_2022_CVPR}
{\sc Yang, F., Wu, K., Zhang, S., Jiang, G., Liu, Y., Zheng, F., Zhang, W.,
  Wang, C., and Zeng, L.}
\newblock Class-aware contrastive semi-supervised learning.
\newblock In {\em Proceedings of the IEEE/CVF Conference on Computer Vision and
  Pattern Recognition (CVPR)\/} (June 2022), pp.~14421--14430.

\bibitem{DBLP:conf/iccv/YueZZSKG19}
{\sc Yue, X., Zhang, Y., Zhao, S., Sangiovanni{-}Vincentelli, A.~L., Keutzer,
  K., and Gong, B.}
\newblock Domain randomization and pyramid consistency: Simulation-to-real
  generalization without accessing target domain data.
\newblock In {\em International Conference on Computer Vision (ICCV)\/} (2019),
  pp.~2100--2110.

\bibitem{DBLP:conf/nips/ZhangWHWWOS21}
{\sc Zhang, B., Wang, Y., Hou, W., Wu, H., Wang, J., Okumura, M., and
  Shinozaki, T.}
\newblock Flexmatch: Boosting semi-supervised learning with curriculum pseudo
  labeling.
\newblock In {\em Advances in Neural Information Processing Systems
  (NeurIPS)\/} (2021), pp.~18408--18419.

\bibitem{DBLP:journals/pr/ZhangQSG22}
{\sc Zhang, J., Qi, L., Shi, Y., and Gao, Y.}
\newblock Generalizable model-agnostic semantic segmentation via
  target-specific normalization.
\newblock {\em Pattern Recognition (PR) 122\/} (2022), 108292.

\bibitem{zhao2022balanced}
{\sc Zhao, J., Liu, X., and Zhao, W.}
\newblock Balanced and accurate pseudo-labels for semi-supervised image
  classification.
\newblock {\em ACM Transactions on Multimidia Computing Communications and
  Applications (TOMM)\/} (2022).

\bibitem{DBLP:conf/nips/ZhaoGLFT20}
{\sc Zhao, S., Gong, M., Liu, T., Fu, H., and Tao, D.}
\newblock Domain generalization via entropy regularization.
\newblock In {\em Advances in Neural Information Processing Systems
  (NeurIPS)\/} (2020).

\bibitem{Zhao_2022_CVPR}
{\sc Zhao, Z., Zhou, L., Duan, Y., Wang, L., Qi, L., and Shi, Y.}
\newblock Dc-ssl: Addressing mismatched class distribution in semi-supervised
  learning.
\newblock In {\em IEEE Conference on Computer Vision and Pattern Recognition
  (CVPR)\/} (2022), pp.~9757--9765.

\bibitem{Zheng_2022_CVPR}
{\sc Zheng, M., You, S., Huang, L., Wang, F., Qian, C., and Xu, C.}
\newblock Simmatch: Semi-supervised learning with similarity matching.
\newblock In {\em IEEE Conference on Computer Vision and Pattern Recognition
  (CVPR)\/} (2022), pp.~14471--14481.

\bibitem{zhou2021domain}
{\sc Zhou, K., Liu, Z., Qiao, Y., Xiang, T., and Loy, C.~C.}
\newblock Domain generalization: A survey.
\newblock {\em IEEE Transactions on Pattern Analysis and Machine Intelligence
  (TPAMI)\/} (2022).

\bibitem{DBLP:journals/corr/abs-2106-00592}
{\sc Zhou, K., Loy, C.~C., and Liu, Z.}
\newblock Semi-supervised domain generalization with stochastic stylematch.
\newblock {\em arXiv preprint arXiv:2106.00592\/} (2021).

\bibitem{DBLP:conf/aaai/ZhouYHX20}
{\sc Zhou, K., Yang, Y., Hospedales, T.~M., and Xiang, T.}
\newblock Deep domain-adversarial image generation for domain generalisation.
\newblock In {\em AAAI Conference on Artificial Intelligence (AAAI)\/} (2020),
  pp.~13025--13032.

\bibitem{DBLP:journals/tip/ZhouYQX21}
{\sc Zhou, K., Yang, Y., Qiao, Y., and Xiang, T.}
\newblock Domain adaptive ensemble learning.
\newblock {\em IEEE Transactions Image Process (TIP) 30\/} (2021), 8008--8018.

\bibitem{Zhou_2022_CVPR}
{\sc Zhou, Z., Qi, L., Yang, X., Ni, D., and Shi, Y.}
\newblock Generalizable cross-modality medical image segmentation via style
  augmentation and dual normalization.
\newblock In {\em Proceedings of the IEEE/CVF Conference on Computer Vision and
  Pattern Recognition (CVPR)\/} (June 2022), pp.~20856--20865.

\bibitem{DBLP:conf/iccv/ZhuPIE17}
{\sc Zhu, J., Park, T., Isola, P., and Efros, A.~A.}
\newblock Unpaired image-to-image translation using cycle-consistent
  adversarial networks.
\newblock In {\em International Conference on Computer Vision (ICCV)\/} (2017),
  pp.~2242--2251.

\end{thebibliography}
}
\end{document}